\documentclass[lettersize,journal]{IEEEtran}
\usepackage{amsmath,amsfonts}
\usepackage{algorithmic}
\usepackage{algorithm}
\usepackage{array}
\usepackage[caption=false,font=normalsize,labelfont=sf,textfont=sf]{subfig}
\usepackage{textcomp}
\usepackage{stfloats}
\usepackage{url}
\usepackage{verbatim}
\usepackage{graphicx}
\usepackage{cite}
\usepackage{threeparttable}
\usepackage{tikz}
\usetikzlibrary{positioning}
\usepackage{subcaption}
\usepackage{multirow}
\usepackage{tabularx}
\usepackage{booktabs}
\usepackage{rotating}
\usepackage{hyperref}
\hypersetup{
    colorlinks=true,
    linkcolor=blue,
    citecolor=blue,
    urlcolor=blue
}
\usepackage{orcidlink}
\begin{document}

\title{High-Resolution Flood Mapping With Sentinel-1 and Sentinel-2 via Misalignment-Robust Cross-Sensor Learning and Generative Despeckling}

\author{David~Ma\,\orcidlink{0000-0002-0203-6539}, Jeremy~Feinstein\,\orcidlink{0009-0009-5362-6860}, Shreya~Pandit\,\orcidlink{0009-0002-8567-1876}, Arkaprabha~Ganguli\,\orcidlink{0000-0002-5588-2157}, and Eugene~Yan\,\orcidlink{0000-0002-7112-7397}%
\thanks{This material is based upon work supported by Laboratory Directed Research and Development (LDRD) funding from Argonne National Laboratory, provided by the Director, Office of Science, of the U.S. Department of Energy under Contract No. DE-AC02-06CH11357.}%
\thanks{David Ma, Jeremy Feinstein, Shreya Pandit, Arkaprabha Ganguli, and Eugene Yan are with Argonne National Laboratory, Lemont, IL 60439 USA (e-mail: dma@anl.gov; jfeinstein@anl.gov; eyan@anl.gov).}%
\thanks{David Ma is also with the University of Chicago, Chicago, IL 60637 USA.}%
\thanks{Shreya Pandit is also with Ohlone College, Fremont, CA 94539 USA.}%
}


\maketitle

\begin{abstract}
Reliable high-resolution flood extent mapping from satellite imagery remains constrained by limited data fidelity and sensor-specific artifacts. Multispectral optical imagery is degraded by clouds, shadows, and urban confounders, while synthetic aperture radar (SAR) imagery is affected by speckle noise and sensor co-registration uncertainty. This work presents an integrated flood mapping framework that jointly addresses these limitations through curated datasets and novel learning strategies. We introduce a new Sentinel-2 (S2) and Sentinel-1 (S1) dataset covering the contiguous United States, featuring pixel-accurate 10 m water masks with emphasis on challenging weather conditions and urban environments that are underrepresented in existing benchmarks. High-quality S2 annotations are manually produced using rigorous geospatial labeling protocols and transferred to SAR imagery through weakly labeled temporally coincident acquisitions. To address SAR-specific artifacts, a shift-invariant loss function is employed to tolerate residual geolocation uncertainty between SAR imagery and optical-derived labels, and a Conditional Variational Autoencoder (CVAE) is trained on multitemporal SAR composites to suppress speckle while preserving flood-relevant spatial structure. Experiments using UNet and UNet++ architectures demonstrate strong multispectral performance (AUPRC up to 0.956) and statistically significant improvements in SAR flood mapping when using shift-invariant loss and CVAE-based despeckling compared to classical filters. These results underscore the importance of dataset fidelity, misalignment-robust training, and demonstrate the viability of generative despeckling for operational flood mapping.
\end{abstract}

\begin{IEEEkeywords}
remote sensing, Sentinel-2, Sentinel-1, segmentation, flood mapping, synthetic aperture radar, despeckling, deep learning, convolutional neural network, generative deep learning, conditional variational autoencoder.
\end{IEEEkeywords}

\section{Introduction}
Flooding contributes a significant economic burden to the United States each year. The Joint Economic Committee estimates that the total cost of flooding in the US ranges from \$179.8 to \$496.0 billion annually, equivalent to 1-2\% of the US GDP \cite{JEC2024FloodingCosts}. Moreover, flood risk and prevalence are only expected to worsen with the impacts of climate change \cite{warmfloodrisk, warmfloodrisk2}. The severe socioeconomic damage and increasing risks of floods highlight the need to invest in flood emergency relief, supported by the development of comprehensive flood maps for extent monitoring, damage assessment, and resource administration. The information provided by flood maps helps emergency responders, field operators, utility companies, and local governments respond quickly and effectively during crises. Flood mapping tools can also be used to prepare for future disasters, such as improving hydrologic models for extreme-event simulations. Hence, the development of more capable flood mapping methodologies is critical to reducing the short- and long-term socioeconomic impacts of flooding.

In recent years, flood mapping through remote sensing has been widely adopted over traditional approaches such as field surveys or physics-based numerical models due to its lower cost, rapid deployment, broad spatial coverage, and scalability \cite{remotefloods, numericalswe, japanfloods}. Remote sensing draws on a diverse ecosystem of satellite missions to collect global observations of surface flood events. These observations are obtained in two primary sensing modalities: multispectral optical imagery and synthetic aperture radar (SAR). Multispectral imagery measures surface reflectance across visible and infrared bands, enabling direct identification of open water based on its spectral characteristics. In contrast, SAR is a side-looking radar sensor that transmits microwave pulses and measures the returned backscatter signal, providing observations of surface water independent of cloud cover or solar illumination.

To assess flood extent in multispectral and SAR imagery, semi-automated to fully automated techniques are typically employed, from water index and backscatter thresholding \cite{sfmi, waterindices, awei, sarthreshold}, to change detection \cite{vekaria, sarchangedetection, sarparathreshold}, to more data-driven deep learning (convolutional or transformer-based) methods \cite{bereczky, deepsarflood, sarglobalfloods}. Among these approaches, deep learning methods have demonstrated comparatively stronger performance by automatically learning spectral, backscatter, and spatial feature representations directly from large-scale remote sensing datasets, enabling improved discrimination of flooded and non-flooded surfaces across diverse land cover types and imaging conditions \cite{bereczky, bentivoglio}.

However, despite advances in deep learning-based flood mapping, reliable high-resolution flood extent estimation remains limited by sensor-specific challenges that continue to constrain model robustness and generalization. In the multispectral modality, clouds, shadows, and urban areas are among several classes that remain difficult for accurate water detection. Thick clouds can obstruct the visibility of surface water and prevent water extraction \cite{shastrycloud, tarpanelli}, while thin cirrus clouds partially contaminate the underlying spectral signal, degrading prediction performance \cite{thincirrusremoval}. Cloud shadows and topographic shadows can also exhibit water-like reflectance, leading to increased false positives in predicted flood extents \cite{shadowremove1, shadowremove2, shadowfalsepos}. A related challenge arises in urban areas, where highly reflective or dark objects such as asphalt pavements and rooftops frequently take on similar signatures as water \cite{s2urbanpixels}. Combined, these classes often cause spurious and noisy model predictions in the multispectral domain.

On the other hand, the SAR modality is primarily affected by the phenomenon of speckle noise, a coherent interference of signals from multiple elementary scatterers that degrades image quality and complicates precise, high-resolution segmentation relative to multispectral imagery \cite{sarspeckle}. The side-viewing geometry presents a secondary set of challenges. SAR must be converted from slant range to ground range through geometric correction for co-registration with nadir-view optical imagery \cite{sargeometrydist, sars2positioning}. However, image co-registration is still an area of active research \cite{coregistration1, coregistration2, coregistration3, coregistration4}, with residual distortions and sensor to sensor spatial misalignments adversely impacting cross-modal model training and performance.

These challenges can, in principle, be mitigated in deep learning models through dataset curation that emphasizes geospatial diversity, high-fidelity annotations, and correction of SAR noise and cross-sensor misalignment. However, existing publicly available flood mapping datasets are not optimized for these requirements, making training data quality the primary performance bottleneck rather than model architecture \cite{bereczky}. For instance, many widely used Sentinel-2 (S2) and Sentinel-1 (S1) datasets employ semi-automated thresholding methods that compromise annotation fidelity. Sen1Floods11 applies Otsu thresholding \cite{otsu} for S1 and NDWI/MNDWI thresholding for S2 data \cite{sen1flood11}. These thresholded labels are prone to misclassification under speckle, cloud cover, shadows, and complex urban land cover. Other datasets avoid simple thresholding, but instead rely on weak labels that have not undergone sufficient quality control. Datasets such as WorldFloods \cite{worldfloods}, Sen12-Flood \cite{sen12flood}, and OMBRIA \cite{ombria}, among others \cite{Notarangelo}, use Copernicus Emergency Management Services (CEMS) flood masks derived from an ensemble of SAR-based algorithms. These labels are known to suffer from temporal misalignment between imagery and masks, coarse delineation of flood boundaries, and limited manual refinement \cite{s1s2water}. Furthermore, many public datasets like S2S1-Water lack sufficient representation of difficult but operationally critical scenes and conditions, including cloud cover, haze, and dense urban environments, which significantly limits model generalization \cite{amitrano}.

Addressing current dataset limitations requires pixel-level annotation fidelity at the native resolution of sensors while spanning a diverse range of geographic regions, land cover types, and imaging conditions. SAR datasets must also explicitly confront residual speckle noise and cross-sensor misregistration to enable reliable, high-resolution water delineation and effective cross-modal model training. Without such datasets, deep learning models remain constrained in their ability to learn fine-grained spatial structure and to generalize across heterogeneous, real-world flood scenarios.

In response to these needs, this work makes the following two primary contributions:

\begin{enumerate}
    \item Introduces a high-quality S2 and S1 dataset of flood and non-flood events covering the contiguous United States (CONUS), featuring pixel-accurate 10m water masks and a focus on challenging weather conditions and urban scenes that are underrepresented in existing open datasets.
    \item Proposes novel training and inference paradigms, including a shift-invariant loss to address cross-sensor geolocation uncertainty between SAR and multispectral images, and a generative denoising approach using Conditional Variational Autoencoders (CVAE) to suppress SAR speckle and improve SAR image quality for downstream flood mapping.
\end{enumerate}

\section{Data}
\label{section:data}

In this study, S2 and S1 observations from the European Space Agency (ESA) were used to generate historical flood maps owing to their open-access data policy and comprehensive coverage across the CONUS.

Three primary datasets were generated for training models within the high-resolution flood mapping pipeline:
\begin{enumerate}
    \item \textbf{S2 flood segmentation dataset}: Curated set of 146 multispectral flood images (4 km $\times$ 4 km) across CONUS with manually labeled 10m water masks (Sec. \ref{section:multispectral}, \ref{sampling}, \ref{section:topographyroads}, \ref{section:manual-annotation}, \ref{section:split}, \ref{section:patching}, \ref{section:trainingdata}).
    \item \textbf{S1 flood segmentation dataset}: Curated set of 7,846 SAR flood images (4 km $\times$ 4 km) across CONUS with weakly labeled 10m water masks (Sec. \ref{section:sar}, \ref{sampling}, \ref{section:topographyroads}, \ref{section:weaklabeling}, \ref{section:split}, \ref{section:patching}, \ref{section:trainingdata}).
    \item \textbf{S1 multitemporal composite dataset}: Dataset of 38,252 multitemporal stacks of 7 consecutive non-flood related SAR images (4 km $\times$ 4 km) across the CONUS, split into pre-training and fine-tuning subsets (Sec. \ref{section:sar}, \ref{section:multitemporal}, \ref{section:split}, \ref{section:patching}, \ref{section:trainingdata}). This dataset is used exclusively to train the CVAE-based SAR despeckling model and does not contain water masks.
\end{enumerate}

The procedures for generating the datasets, along with the supplementary geophysical layers used during processing and model training, are described in detail below.

\subsection{Multispectral Data}
\label{section:multispectral}
The Sentinel-2 mission provides global \(10 \, \text{m}\) multispectral data from 2015 to 2026 \cite{sentinel2}. The visible (red, green, and blue), near-infrared (NIR), and two short wave infrared (SWIR1 and SWIR2) bands from S2 were used along with four water indices derived from the surface reflectance. Water indices are scores computed from combinations of spectral bands that quantify the relative likelihood of water presence. These include the Normalized Difference Water Index (NDWI), the Modern Normalized Difference Water Index (MNDWI), and the Automated Water Extraction Index with shadow (AWEI\textsubscript{sh}) and without shadow (AWEI\textsubscript{nsh}) \cite{ndwi, mndwi, awei}. The water indices are defined by the following equations, with band names corresponding to band reflectance:
\begin{equation} \label{eq:ndwi}
\text{NDWI} = \frac{\text{Green} - \text{NIR}}{\text{Green} + \text{NIR}}
\end{equation}

\begin{equation} \label{eq:mndwi}
\text{MNDWI} = \frac{\text{Green} - \text{SWIR1}}{\text{Green} + \text{SWIR1}}
\end{equation}

\begin{multline}\label{eq:aweinsh}
\text{AWEI}_{\text{nsh}} = 4 \times (\text{Green} - \text{SWIR1}) \\
- (0.25 \times \text{NIR} + 2.75 \times \text{SWIR2})
\end{multline}

\begin{multline} \label{eq:aweish}
\text{AWEI}_{\text{sh}} = \text{Blue} + 2.5 \times \text{Green} \\
- 1.5 \times (\text{NIR} + \text{SWIR1}) - 0.25 \times \text{SWIR2}
\end{multline}

NDWI and MNDWI produce normalized values in the range $[-1, 1]$, while AWEI\textsubscript{sh} and AWEI\textsubscript{nsh} produce unnormalized continuous values that can extend beyond this range. Each water index differs in ability to separate water from specific land cover types \cite{waterindices}. For instance, the NDWI helps distinguish and contrast the moisture content of water bodies from surrounding soil and vegetation, but struggles to distinguish mixed water pixels from urban pixels \cite{ndwi}. Meanwhile, the AWEI\textsubscript{nsh} index can effectively separate mixed water from urban pixels but has a tendency to confuse mountain shadow with water. Given the variability in their coverage, all four indices were included rather than only one to combine the strengths of each.

S2 Level-2A (L2A) products were accessed via the Microsoft Planetary Computer and the Copernicus Data Space Ecosystem L2A catalog. Band values (stored as digital numbers; DN) were converted to surface reflectances. To ensure consistency across the S2 processing baseline update effective 25th January 2022, we applied the documented additive offset ($0$ DN prior to this date; $-1000$ DN on and after) indicated in the product metadata and S2 documentation \cite{sentinel2}.

Following correction, the water indices (NDWI, MNDWI, AWEI) were computed from surface reflectance values.

\subsection{Synthetic Aperture Data}
\label{section:sar}
The Sentinel-1 mission provides global SAR data in multiple product forms \cite{sentinel1}. Single Look Complex (SLC) products contain high-resolution amplitude and phase information in slant-range geometry, while Ground Range Detected (GRD) products are multi-looked to reduce speckle noise and projected to ground range geometry, yielding backscatter intensity at \(10 \, \text{m}\) resolution. In SLC, speckle noise can be modeled mathematically as a sum of contributions of $N$ multiple independent scatterers with complex amplitude

\begin{equation}
    Z = \sum^{N}_{k=1} a_k e^{i \phi_k}
\end{equation}

where $a_k$ and $\phi_k$ represent the amplitude and phase of the $k$th scatterer \cite{goodmanspeckle}. Under the fully developed speckle assumption, the observed SAR intensity $Y$ in a GRD image can be expressed using a multiplicative speckle model,

\begin{equation}
\label{eq:mult}
Y = X \cdot S
\end{equation}

where $X$ represents the underlying reflectivity and 
$S$ models speckle noise in a Gamma distribution with probability density function

\begin{equation}
p(i) =
\frac{L^L}{\Gamma(L)}
i^{L-1} e^{-Li},
\quad i \ge 0
\end{equation}

where $L \geq 1$ denotes the independent number of looks and $\Gamma(\cdot)$ is the Gamma function. When converted to decibels (logarithmic scale), speckle noise contribution in Equation \ref{eq:mult} becomes additive.

For the purposes of the study, GRD products with Interferometric Wideswathe (IW) acquisition mode were used due to their extensive land coverage across the states. IW mode uses Terrain Observation with Progressive Scanning SAR (TOPSAR) which collects and stitches three overlapping subswathes into a much wider swathe. VV and VH polarizations were chosen based on their relevance in flood extent modeling in previous studies \cite{vekaria, bereczky}. SAR products were downloaded from the Sentinel-1 Radiometrically Terrain Corrected (RTC) dataset provided by Microsoft Planetary Computer. The additional radiometric terrain correction reduces topography induced artifacts and improves the usefulness of backscatter intensities for land cover detection. The downloaded products were converted to dB scale and normalized with training mean and standard deviation.

\subsection{Spatiotemporal Sampling By Precipitation}
\label{sampling}

\begin{figure*}[!t]
    \centering
    \includegraphics[width=\textwidth]{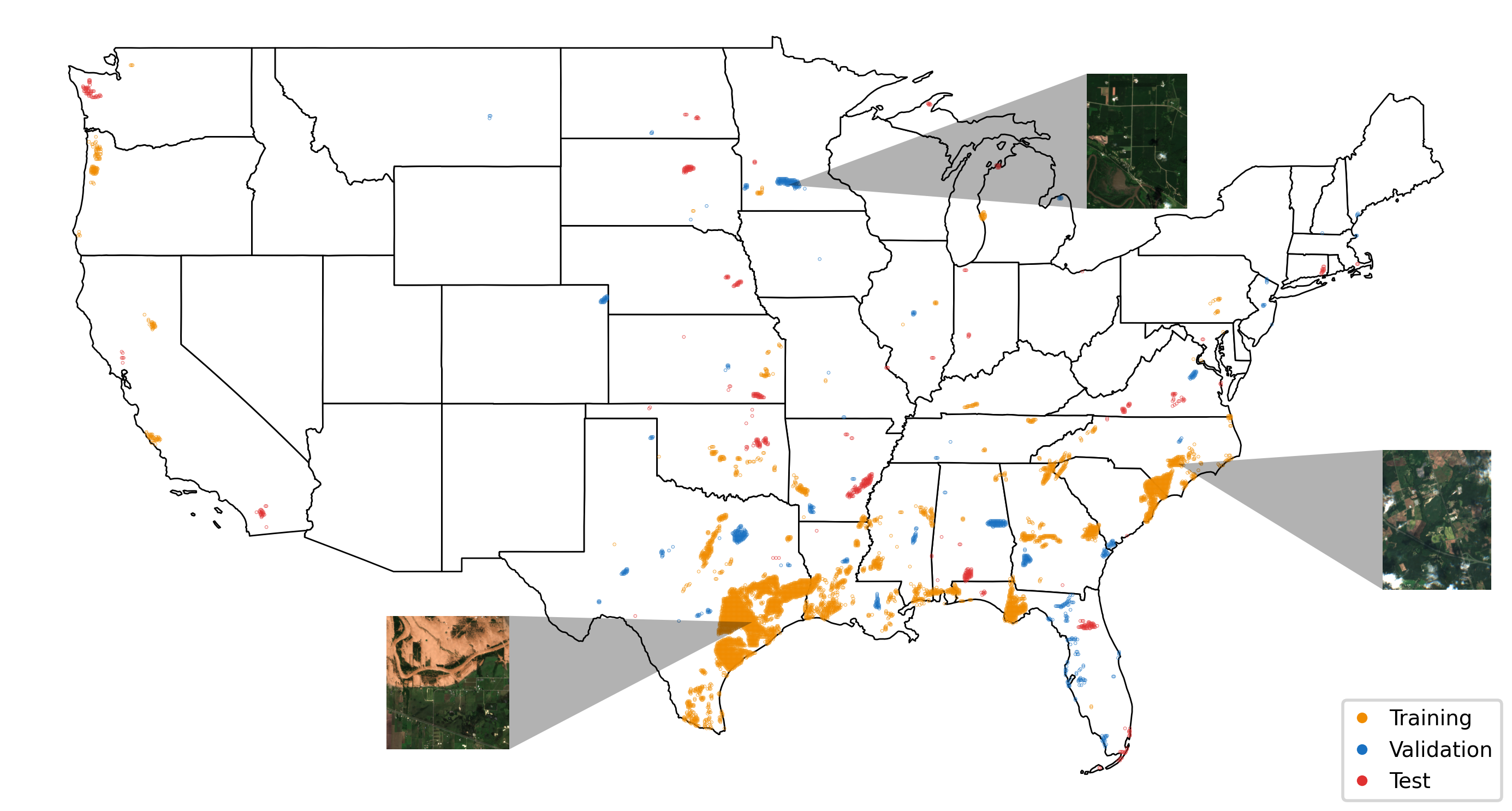}
    \caption{The S1 SAR dataset tile locations sampled within the CONUS using spatiotemporal precipitation filtering (N=7846). Each dot represents a 4km by 4km PRISM cell, with dense clustering due to the localized nature of storm events. Examples of the temporally coincident S2 product True Color Image (TCI) are displayed.}
    \label{fig:sardataset}
\end{figure*}

A two-pronged methodology was used to sample S2 multispectral and S1 SAR surface flood imagery across the CONUS, with emphasis placed on areas in the Midwest (Illinois, Wisconsin and Indiana) and South (Texas, Florida, and Mississippi). Sampling used two independent strategies: extreme precipitation thresholding and manual selection.

Because heavy or prolonged rainfall can cause and intensify many types of disastrous flood events \cite{rainelastic, riverflood, flashflood, coastal, ilfloods}, satellite images taken of areas after receiving high cumulative precipitation were sampled to target flooded pixels. The extreme precipitation sampling was conducted by thresholding daily precipitation in the 4km PRISM Climate Group (PRISM) dataset \cite{prism} and cross-referencing with S1 and S2 products.  A threshold of 150mm (5.9in) was used to balance flood pixel availability and event extremeness. For multispectral data, all S2 products in the timespan from 4 days before to 6 days after these events were downloaded (unless no post-precipitation data existed). For SAR data, pairs of S1 and S2 products captured within 24 hours of each other were downloaded if they were taken within 5 days after these events. The S2 and S1 imagery downloaded for each PRISM cell were each cropped to the 4km x 4km bounding box, resulting in roughly 400 by 400 pixel images.

The second method of manual sampling was used to capture flood pixels when they existed downstream of precipitation (i.e., they would be excluded from the threshold-based sampling explained above) or were not caused by precipitation. Using the Copernicus browser tool, S2 granules were visualized as a True Color Image (TCI), and any cells with apparent flood imagery were subsequently added to the dataset. This procedure was performed for areas adjacent to PRISM cells identified in the threshold-based sampling approach and for major flood events found through  a review of newspaper articles and government flood records in Texas, Illinois, and surrounding states. Notably, the PRISM grid cell was also used here to crop the manually selected flood images because of its role as the unit of discretization. The 400 by 400 pixel satellite images of PRISM grid cells are referred to throughout the rest of the paper as ``tiles''.

As a result of both extreme precipitation and manual sampling, thousands of tiles were collected and combined as part of the multispectral and SAR dataset shown in Table \ref{table:dataset-summary}.

\subsection{Topography and Roads}
\label{section:topographyroads}
Cloud, terrain shadows, or smooth man-made surfaces (e.g., asphalt roads or runways) are often misclassified for water algorithmically due to similar signatures in optical and radar imagery \cite{waterindices, sardemmap}. The following supplementary layers are added to each sampled tile in \ref{sampling} to provide geophysical context and guide the models toward more plausible flood detection across such cases:
\begin{itemize}
    \item Digital elevation models (DEM) with \(10 \, \text{m}\) resolution from the National Elevation Dataset via the USDA NRCS Geospatial Data Gateway \cite{usda_dem}.
    \item Permanent waterbodies and flowlines from the National Hydrography Dataset High Resolution (NHDPlus HR)  \cite{nhd}.
    \item Regional roads from the US Census TIGER/Line Roads dataset \cite{tiger}.
\end{itemize}

\subsection{Cloud Coverage and Multispectral Limitations}
\label{cloudcover}
A major obstacle in the data collection of multispectral flood imagery is the obstruction from atmospheric cloud cover, with one study \cite{tarpanelli} reporting that only 28\% of flood events were observable by S2 due to cloud coverage. This results in a large percentage of S2 data sampled from \ref{sampling} lacking any valuable flood pixels. A threshold was applied to exclude candidate tiles based on the high percentage of cloud cover generated from the L2A scene classification algorithm \cite{sentinel2}.

A secondary challenge exists where the surface contains cloud shadows, haze, or urban scenes, affecting the ability to make accurate and reliable predictions on flood extent using multispectral imagery \cite{shadowremove1, shadowremove2, awei}. To improve model robustness in these regimes, high proportions of labeled water pixels under thin cirrus or cloud shadows, and non-water pixels in urban areas were included in the S2 dataset.

\subsection{Multitemporal Composite}
\label{section:multitemporal}

A common remedy for SAR speckle in remote sensing is to average multiple SAR captures over the same region. This results in a multitemporal composite with smooth homogeneous textures and sharper structural features, at the cost of temporal resolution and consistency. As the closest approximation to speckle free SAR images, multitemporal composites serve as useful targets in training a deep learning model to despeckle. Hence, a dataset of single-to-multitemporal composite SAR image pairs was collected for training a SAR despeckling model.

For each PRISM cell, up to 3 multitemporal stacks of 7 consecutive SAR images (each stack across a maximum of 42 days) were downloaded. The choice of 3 independent time intervals per cell aimed to increase temporal diversity and improve robustness in despeckling scenes across seasons. The multitemporal composite stack size of 7 was determined by empirical judgments of sharpness and image consistency tradeoffs.

The dataset was sampled in two partitions: pre-training and fine-tuning. For the pre-training partition, PRISM cells in the CONUS were selected using random uniform sampling, with cells from the S1 dataset excluded. For the fine-tuning partition, only PRISM cells from the S1 dataset in \ref{sampling} were sampled. The division aimed to ensure that models can be pre-trained to despeckle general SAR scenes across the CONUS before fine-tuning toward flood areas present in SAR model training. Special care was taken to ensure SAR acquisitions in each interval shared the same orbit direction (ascending or descending) to prevent averaging incongruous geometries.

The compositing process used a leave-one-out method, with each noisy SAR image  paired with the composite from the remaining 6. This step was added to prevent bias in the composite image toward speckle in the single SAR image. For each stack of 7 images, 7 pairs of single-to-multitemporal composite images were generated. The data collected is summarized in Table \ref{table:dataset-summary-2}.

\subsection{Manual Annotation}
\label{section:manual-annotation}

S2 and S1 public datasets often lack accurate, truly 10m water masks, limiting the ability of high-resolution flood models in challenging scenes (e.g., narrow streets, dense urban areas, and forest canopies). The quality gap emphasizes the need for robust, policy-driven manual annotations.

Here, manual labeling followed a systematic set of pixel-based labeling instructions that combined various sources of geospatial information. In addition to using color mapped visualizations of S2 products, True Color Image (TCI), NDWI, DEM, SCL, NLCD, slope, roads, waterbodies, and flowlines, protocols were established for labeling flood pixels based on weather conditions, cloud cover, and resolution-based visual ambiguity. The full list of policies is provided in Table \ref{table:labeling-policies}.

Focusing on quality water masks comes with the inherent tradeoff of labeling speed. In the first stages of ground truthing the S2 dataset, each water pixel in the tile needed to be selected by hand. But as throughput increases later on with machine-assisted labels, this cost becomes negligible. Once the dataset reached sufficient size for model training, the preliminary models produced machine labels with similar pixel precision, requiring only slight manual corrections to be incorporated into the dataset. Thus, the initial time consuming hand labeling process turned into a fast semi-automated approach with iterative machine labeling and correction to build out a dataset of S2 water masks accurate to the native 10m image resolution.

The final S2 dataset consisted of 146 manually annotated event tiles.

\begin{table*}[!t]
\centering
\caption{Labeling policy instructions and justifications.}
\label{table:labeling-policies}
\begin{tabular}{p{0.05\linewidth} p{0.35\linewidth} p{0.40\linewidth}}
\hline
\textbf{\#} & \textbf{Policy instruction} & \textbf{Justification} \\
\hline
1 &
Treat permanent water, non-permanent water, and flood water as a single positive class. &
Simplifies water labeling objectives and avoids inconsistent class boundaries. \\

2 &
For each tile, review color-mapped S2 products and supplementary channels: TCI, NDWI, DEM, SCL, NLCD, slope, roads, waterbodies, flowlines. &
Provides evidence for water presence and important context (e.g., terrain, land cover, mapped hydrography) to reduce omissions and commission errors under variable surface and atmospheric conditions. \\

3 &
Overlay georeferenced rasters with high-resolution Google Satellite imagery in QGIS during interpretation. &
Enables pixel-level disambiguation by leveraging finer spatial detail, improving boundary placement and reducing false positives from urban features. \\

4 &
Edit the water mask in GIMP using a pencil tool while reviewing the QGIS overlays. &
Allows direct per-pixel additions/removals to match the native 10\,m grid, supporting high-fidelity masks and precise shoreline/flood-edge adjustments. \\

5 &
Prioritize TCI, NDWI, and high-resolution imagery as primary evidence sources during labeling. &
TCI and NDWI provide the strongest optical indicators of water presence, while high-resolution imagery supplies decisive contextual confirmation for borderline pixels (e.g., shadows, dark roofs, asphalt). \\

6 &
Use high-resolution imagery to identify small permanent water bodies (e.g., $\sim$10\,m scale) not captured in NHD-derived waterbody masks. &
Compensates for incompleteness in vector hydrography layers, improving recall for small ponds/ditches and narrow features at the 10m resolution. \\

7 &
Water visible through thin cirrus or haze should be labeled positively, but if the surface is completely obstructed by thick cumulus clouds, it should be labeled as negative. &
Improves robustness of model for water presence prediction under adverse weather conditions (e.g., haze) while also preventing excess false positives from cloud pixels. \\

8 &
When available for specific events, incorporate high-resolution aerial flood imagery (e.g., NOAA imagery for Hurricane Harvey tiles). &
Addresses the limitation that high resolution satellite imagery are composited from non-flood dates; event-specific aerial imagery can directly validate flood extent with high granularity. \\

9 &
Early-stage masks may require exhaustive manual labeling; later stages should use machine-assisted labels with additional manual correction. &
The speed-quality tradeoff front-loads labeling time to establish a reliable training set; once models achieve sufficient performance, iterative prediction using machine assisted labels accelerates throughput while maintaining pixel-level precision. \\
\hline
\end{tabular}
\end{table*}

\subsection{Weak Labeling}
\label{section:weaklabeling}
The interpretability and spatial resolution of S2 imagery were leveraged to provide high-fidelity labels for the SAR dataset. For each S1 product in the SAR dataset, annotations from the temporally coincident (within 24 hours) S2 product were used as an estimate of SAR water extent. This enabled the transfer of high quality labels from visual imagery to the SAR domain while avoiding the pitfalls of SAR-specific thresholding methods.

In order to develop a large scale SAR dataset, a weak labeling approach was employed to large numbers of SAR tiles where temporally coincident S2 manual annotation labels were unavailable. Machine labels for each SAR tile were generated from the best performing tuned and benchmarked S2 multispectral model. Each tile and its weak label was visually inspected for dataset quality prior to inclusion. Weak labels with artifacts due to missing data and low accuracy labels generated from geographies outside of the S2 multispectral dataset (e.g., mountainous and snowy regions) were removed. Tiles containing solely water or covered in overly dense haze were removed. Out of 10115 sampled SAR tiles, 14.7\% were filtered out through manual inspection of label quality. The final curated SAR dataset consisted of 7846 event tiles total, with 61 manually annotated labels and 7785 machine annotated labels.

\subsection{Train, Validation, and Test Split}
\label{section:split}
The S2 multispectral dataset tiles were split into train, validation, and test using a manual 80/10/10 split with stratification by geographic region, scene type, weather condition, and flood presence to reduce training and evaluation bias. As tiles were spatially distributed around Texas and Illinois, geographic stratification was done via membership in the two clusters. To stratify scene type, each tile was categorized as urban (suburban or city) or non-urban. To stratify weather condition, each tile was categorized as containing clear-weather, haze, clouds, or cloud shadows. Lastly, each split was allocated proportionate numbers of flooded tiles. Care was taken to reduce data leakage from correlated weather and solar illumination conditions by placing all tiles from the same S2 product acquisition in the same split.

For the S1 SAR dataset, splitting was done by grouping tiles into shared temporal and geographic clusters before randomly assigning clusters to train, validation, and test in an 80/10/10 split. Given the large quantity of tiles, automatic clustering was performed using the following steps:

\begin{enumerate}
    \item Group tiles across different datatake dates that share the same PRISM cell.
    \item Group tiles within 5 cells of each other in the PRISM grid by manhattan distance.
    \item Merge groups that overlap in datatake dates.
\end{enumerate}

The automatic grouping resulted in 104 clusters of tiles separate in geographic area and imagery acquisition dates. Clusters were then distributed randomly among training, validation, and test sets with stratification of flood presence. The 22 clusters containing flood imagery were distributed such that 50\% of clusters in each of the validation and test sets contained flood imagery. Remaining flood clusters were allocated to the training set.

Additional steps were taken to reduce data leakage in the S1 dataset. A potential source is weather condition. Despite S1 being an all-weather instrument, the C-band instrument has been found to be attenuated by precipitation in the atmosphere \cite{sarweather}. Thus, tiles from the same datatake may share weather-related correlations. Spatial proximity of tiles may also cause leakage. Due to the nature of IW TOPSAR acquisition and processing, tiles sharing borders may contain overlap regions in the IW subswathe mosaicking, resulting in backscatter noise leaking across tiles. Hence, to reduce the risk of data leakage from these two sources, SAR tiles cropped from the same datatake were kept in the same split.

For the multitemporal composite SAR dataset, the pre-training partition followed a random 80/10/10 split. Tiles were first clustered based on cell membership followed by proximity. The clusters were subsequently split into train, val, test sets by random assignment in blocks of 1 degree latitude and longitude. As for the fine-tuning partition, the S1 dataset split was used. The fine-tuning test set was held out for reporting final despeckling metrics to prevent information leaking into downstream SAR flood mapping model training.

\subsection{Dataset Patching}
\label{section:patching}

Two methods for converting $400 \times 400$ pixel tiles into smaller $64 \times 64$ patches were evaluated: strided patch sampling and random patch sampling. Strided patch sampling uses a sliding window horizontally and vertically to extract patches. The stride can be adjusted to increase overlap, ensuring that pixels appearing near patch boundaries are also observed near patch centers. On the other hand, random patch sampling uniformly samples the set of interior patches within a tile until a number of desired patches is reached. The stochastic sampling creates arbitrary patch overlaps that have a similar function to strided patch overlap.

To optimize patching method for model performance, a preliminary experiment was conducted on the S2 dataset. Each dataset patching method was applied to the training set to create $64 \times 64$ patches while keeping the model hyperparameters fixed. For the strided sampling method, strides of 64, 48, 40, 32, 24, 16 were tested. For the random sampling method, patch counts per tile of 100, 250, 500, 750, 1000 were tested. The model was trained on each configuration and then benchmarked on a held out test set. The benchmarks showed that strided patch sampling across all strides outperformed random patch sampling across all patch counts. Based on the optimal benchmark performance, strided $64 \times 64$ patch sampling with a stride of $16$ was chosen for the S2 dataset.

For the S1 flood segmentation dataset, tiles were patched into slightly wider \(68 \times 68\) pixel patches to allow for the shift invariant loss computation as outlined in \ref{section:shift-invariant}, but the same strided patch sampling method was applied. As the dataset is much larger in size, a stride of $68$ was chosen to reduce the total dataset size for memory efficiency.

\subsection{Training Data}
\label{section:trainingdata}
Two primary datasets of multispectral and SAR patches were preprocessed for water detection. The information for each is described in Table \ref{table:dataset-summary}. Both datasets contain DEM, slope Y, slope X, waterbody, roads, and flowlines. The TCI, SCL, and NLCD were also included for model evaluation purposes.

\begin{table*}[htbp]
    \centering
    \caption{Summary of multispectral and SAR flood segmentation datasets.}
    \label{table:dataset-summary}
    \begin{tabularx}{0.8\textwidth}{l >{\raggedright\arraybackslash}X >{\raggedright\arraybackslash}X}
        \toprule
        \textbf{Dataset} & \textbf{Multispectral (S2)} & \textbf{SAR (S1)} \\
        \midrule
        \textbf{Total tiles} & 146 & 7,846 \\
        \addlinespace[2pt]
        \textbf{Ground truth source} & Manual annotation & Water mask generated by trained multispectral model on proximal S2 captures. 61 manual annotations from the S2 dataset were also reused. \\
        \addlinespace[2pt]
        \textbf{Splits (by tile)} & 80\% training, 10\% validation, 10\% test  & 80\% training, 10\% validation, 10\% test \\
        \addlinespace[2pt]
        \textbf{Input channel count} & 16 & 8 \\
        \addlinespace[2pt]
        \textbf{Input channels} &
       B04 (Red), B03 (Green), B02 (Blue), B08 (NIR), B11 (SWIR1), B12 (SWIR2), NDWI, MNDWI, AWEI\textsubscript{sh}, AWEI\textsubscript{nsh}, DEM, slope Y, slope X, waterbody, roads, flowlines &
        VV, VH, DEM, slope Y, slope X, waterbody, roads, flowlines \\
        \addlinespace[2pt]
        \textbf{Patch size} & $64 \times 64$ & $68 \times 68$ \\
        \addlinespace[2pt]
        \textbf{Stride} & 16 & 68 \\
        \addlinespace[2pt]
        \textbf{Training patches} & 80,929 & 256,795 \\
        \bottomrule
    \end{tabularx}
\end{table*}

In addition, two multitemporal SAR composite datasets were preprocessed for pre-training and fine-tuning of the generative despeckling model. The information is described in Table \ref{table:dataset-summary-2}. The channels include single slice VV, single slice VH, composite VV, and composite VH.

\begin{table*}[!t]
    \centering
    \caption{Summary of multitemporal composite SAR dataset.}
    \label{table:dataset-summary-2}
    \begin{tabularx}{0.8\textwidth}{l >{\raggedright\arraybackslash}X >{\raggedright\arraybackslash}X}
        \toprule
        \textbf{Dataset} & \textbf{Pre-training (Multitemporal)} & \textbf{Fine-tuning (Multitemporal)} \\
        \midrule
        \textbf{Total tiles} & 25,940 & 12,312\\
        \addlinespace[2pt]
        \textbf{Stack size (per tile)} & 7 & 7 \\
        \addlinespace[2pt]
        \textbf{Splits (by tile)} & 80\% training, 10\% validation, 10\% test  & 80\% training, 10\% validation, 10\% test \\
        \addlinespace[2pt]
        \textbf{Input channels} & 4 & 4 \\
        \addlinespace[2pt]
        \textbf{Patch size} & $64 \times 64$ & $64 \times 64$ \\
        \addlinespace[2pt]
        \textbf{Stride} & 64 & 64 \\
        \addlinespace[2pt]
        \textbf{Training patches} & 8,558,459 & 4,567,360 \\
        \bottomrule
    \end{tabularx}
\end{table*}

\section{Methods}

\subsection{Model Architectures}

\subsubsection{Segmentation Models}

\begin{figure*}[t]
\centering

\subfloat[UNet architecture\label{fig:unet}]{
  \begin{tikzpicture}[
    node distance=2cm and 2cm,
    scale=0.7,
    every node/.style={
        circle,
        minimum size=0.1cm,
        font=\scriptsize,
        line width=1.2pt
    }
]

\definecolor{encoderBlue}{RGB}{52, 98, 155}
\definecolor{latentTeal}{RGB}{40, 140, 135}
\definecolor{lossGreen}{RGB}{70, 140, 90}

\tikzset{
    flowarrow/.style={
        ->,
        >=stealth,
        line width=1.3pt
    },
    skiparrow/.style={
        dashed,
        ->,
        >=stealth,
        line width=1.3pt
    }
}

\node (x00) [draw=encoderBlue, fill=encoderBlue!22] at (0,0)    {$X^{0,0}$};
\node (x10) [draw=encoderBlue, fill=encoderBlue!22] at (1,-1.5) {$X^{1,0}$};
\node (x20) [draw=encoderBlue, fill=encoderBlue!22] at (2,-3)   {$X^{2,0}$};
\node (x30) [draw=encoderBlue, fill=encoderBlue!22] at (3,-4.5) {$X^{3,0}$};
\node (x40) [draw=encoderBlue, fill=encoderBlue!22] at (4,-6)   {$X^{4,0}$};

\node (x31) [draw=latentTeal, fill=latentTeal!25]   at (5,-4.5) {$X^{3,1}$};
\node (x21) [draw=latentTeal, fill=latentTeal!25]   at (6,-3)   {$X^{2,1}$};
\node (x11) [draw=latentTeal, fill=latentTeal!25]   at (7,-1.5) {$X^{1,1}$};
\node (x01) [draw=latentTeal, fill=latentTeal!25]   at (8,0)    {$X^{0,1}$};

\node (L) [
    rectangle,
    draw=lossGreen,
    fill=lossGreen!22,
    line width=1.2pt
] at (10,0) {$L$};

\draw[flowarrow] (x00) -- (x10);
\draw[flowarrow] (x10) -- (x20);
\draw[flowarrow] (x20) -- (x30);
\draw[flowarrow] (x30) -- (x40);

\draw[flowarrow] (x40) -- (x31);
\draw[flowarrow] (x31) -- (x21);
\draw[flowarrow] (x21) -- (x11);
\draw[flowarrow] (x11) -- (x01);

\draw[flowarrow] (x01) -- (L);

\draw[skiparrow] (x30) -- (x31);
\draw[skiparrow] (x20) -- (x21);
\draw[skiparrow] (x10) -- (x11);
\draw[skiparrow] (x00) -- (x01);

\end{tikzpicture}
}
\subfloat[UNet++ architecture\label{fig:unetpp}]{
  \begin{tikzpicture}[
    node distance=2cm and 2cm,
    scale=0.7,
    every node/.style={
        circle,
        minimum size=0.3cm,
        font=\scriptsize,
        line width=1.2pt
    }
]

\definecolor{encoderBlue}{RGB}{52, 98, 155}
\definecolor{latentTeal}{RGB}{40, 140, 135}
\definecolor{lossGreen}{RGB}{70, 140, 90}

\tikzset{
    flowarrow/.style={
        ->,
        >=stealth,
        line width=1.3pt
    },
    skiparrow/.style={
        dashed,
        ->,
        >=stealth,
        line width=0.5pt
    }
}

\node (x00) [draw=encoderBlue, fill=encoderBlue!22] at (0,0) {$X^{0,0}$};
\node (x10) [draw=encoderBlue, fill=encoderBlue!22] at (1,-1.5) {$X^{1,0}$};
\node (x20) [draw=encoderBlue, fill=encoderBlue!22] at (2,-3) {$X^{2,0}$};
\node (x30) [draw=encoderBlue, fill=encoderBlue!22] at (3,-4.5) {$X^{3,0}$};
\node (x40) [draw=encoderBlue, fill=encoderBlue!22] at (4,-6) {$X^{4,0}$};

\node (x01) [draw=latentTeal, fill=latentTeal!25] at (2,0) {$X^{0,1}$};
\node (x02) [draw=latentTeal, fill=latentTeal!25] at (4,0) {$X^{0,2}$};
\node (x03) [draw=latentTeal, fill=latentTeal!25] at (6,0) {$X^{0,3}$};
\node (x04) [draw=latentTeal, fill=latentTeal!25] at (8,0) {$X^{0,4}$};

\node (x11) [draw=latentTeal, fill=latentTeal!25] at (3,-1.5) {$X^{1,1}$};
\node (x12) [draw=latentTeal, fill=latentTeal!25] at (5,-1.5) {$X^{1,2}$};
\node (x13) [draw=latentTeal, fill=latentTeal!25] at (7,-1.5) {$X^{1,3}$};

\node (x21) [draw=latentTeal, fill=latentTeal!25] at (4,-3) {$X^{2,1}$};
\node (x22) [draw=latentTeal, fill=latentTeal!25] at (6,-3) {$X^{2,2}$};

\node (x31) [draw=latentTeal, fill=latentTeal!25] at (5,-4.5) {$X^{3,1}$};

\node (L) [
    rectangle,
    draw=lossGreen,
    fill=lossGreen!22,
    line width=1.2pt
] at (10,0) {$L$};

\draw[flowarrow] (x00) -- (x10);
\draw[flowarrow] (x10) -- (x20);
\draw[flowarrow] (x20) -- (x30);
\draw[flowarrow] (x30) -- (x40);

\draw[flowarrow] (x10) -- (x01);
\draw[flowarrow] (x20) -- (x11);
\draw[flowarrow] (x11) -- (x02);

\draw[flowarrow] (x30) -- (x21);
\draw[flowarrow] (x21) -- (x12);
\draw[flowarrow] (x12) -- (x03);

\draw[flowarrow] (x40) -- (x31);
\draw[flowarrow] (x31) -- (x22);
\draw[flowarrow] (x22) -- (x13);
\draw[flowarrow] (x13) -- (x04);

\draw[flowarrow] (x01) -- ++(0,1.5) -- (8,1.5) -| (L);
\draw[flowarrow] (x02) -- ++(0,1.5) -- (8,1.5) -| (L);
\draw[flowarrow] (x03) -- ++(0,1.5) -- (8,1.5) -| (L);
\draw[flowarrow] (x04) -- (L);

\draw[skiparrow] (x00) -- (x01);
\draw[skiparrow] (x01) -- (x02);
\draw[skiparrow] (x02) -- (x03);
\draw[skiparrow] (x03) -- (x04);

\draw[skiparrow] (x10) -- (x11);
\draw[skiparrow] (x11) -- (x12);
\draw[skiparrow] (x12) -- (x13);

\draw[skiparrow] (x20) -- (x21);
\draw[skiparrow] (x21) -- (x22);

\draw[skiparrow] (x30) -- (x31);

\draw[skiparrow] (x20) to[out=25, in=155] (x22);
\draw[skiparrow] (x10) to[out=25, in=155] (x12);
\draw[skiparrow] (x11) to[out=25, in=155] (x13);
\draw[skiparrow] (x10) to[out=25, in=155] (x13);

\draw[skiparrow] (x00) to[out=25, in=155] (x02);
\draw[skiparrow] (x00) to[out=25, in=155] (x03);
\draw[skiparrow] (x00) to[out=25, in=155] (x04);
\draw[skiparrow] (x01) to[out=25, in=155] (x03);
\draw[skiparrow] (x02) to[out=25, in=155] (x04);
\draw[skiparrow] (x01) to[out=25, in=155] (x04);

\end{tikzpicture}
}

\caption{Comparison of UNet and UNet++ architectures. UNet++ introduces dense skip connections between encoder and decoder stages.}
\label{fig:unet_comparison}
\end{figure*}
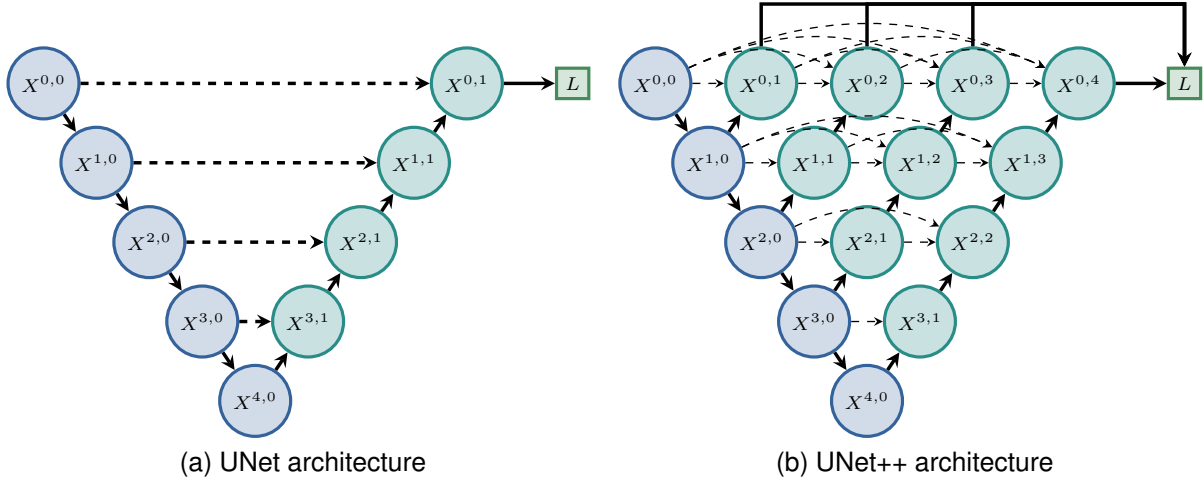

For model architecture, UNet and UNet++ were adopted due to their high performance benchmarks in flood mapping literature in both multispectral and SAR modalities. For example, a study that compared the UNet to other deep learning models such as FCN-8, SegNet and DeepResUNet, found that the UNet achieved better performance in SAR flood mapping in the Yantze River Basin \cite{yangtze}. Similarly, the UNet++ architecture outperformed other state of the art architectures (UNet, DU-Net, R2UNet, AttUNet, NestedUNet, EDSR, and Mask-RCNN) in multispectral segmentation of rivers in Oregon \cite{oregon_rivers}.

The UNet is a fully convolutional neural network following an encoder-decoder layout. The encoder part of the model consists of convolutional layers that downsample the image dimensions while doubling the image channels. The decoder mirrors the encoder by upsampling images back to the original input dimensions and combining learned features to predict the output segmentation mask. Skip connections that send the output of encoder layers directly to distant decoder blocks are used to improve gradient flow and allow recovery of fine-grained details that may have been lost during downsampling \cite{unet}. In contrast to the plain skip connections in the UNet, UNet++ seeks to fuse semantically similar feature maps from the encoder and the decoder to enable more precise and detailed segmentation. UNet++ has demonstrated superior performance when using deep supervision, in which the output of each of the nested UNet structures within the network is averaged for the final segmentation output \cite{nestedunet}. As such, deep supervision was tuned as a hyperparameter.

For the experiments conducted in this study, the UNet and UNet++ were implemented as follows. The UNet and UNet++ use filter sizes $[16, 32, 64, 128, 256]$ for the encoder and decoder blocks. Each block consists of two consecutive 3×3 convolutions with batch normalization and ReLU activation. Downsampling is performed using 2×2 max pooling, while upsampling uses transposed convolutions. Channel wise dropout is applied after the deeper encoder layers for regularization. A 1×1 convolution serves as the output layer for pixel-wise classification.

\begin{figure*}[t]
\centering
\input{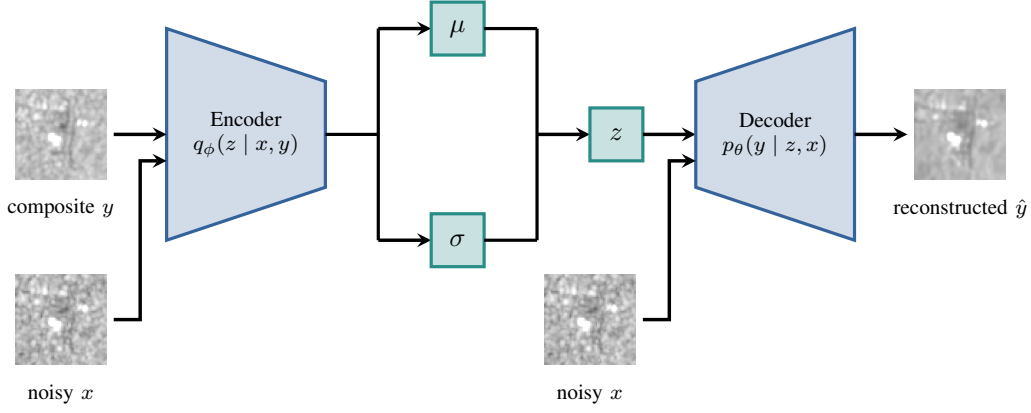}
\caption{CVAE architecture for SAR speckle denoising. The CVAE learns the task of composite SAR patch generation conditioned on noisy SAR patches.}
\label{fig:cvae}
\end{figure*}

\subsubsection{Despeckling Model}

To mitigate the loss of interpretability caused by SAR speckle in downstream tasks (such as surface flood detection), classical speckle filters like Lee, Lee Sigma, Frost, Gamma MAP, Refined Lee, and Enhanced Lee have been developed in remote sensing \cite{sarfilterflood, sarfilterflood2, leefilter, biomassfilter, enhancedlee}. However, these speckle filters typically rely on local statistics and produce oversmoothing that destroys edges and structure critical for image segmentation. In contrast, data-driven methods with supervised, unsupervised, and self-supervised deep learning have shown much more success in SAR despeckling. These include CNNs, GANs, DDPMs, SAR2SAR, Speckle2Void, each demonstrating improved structure preservation and despeckling performance over classical methods \cite{chierchia, gandespeckling, sarddpm, sar2sar, speckle2void}.

The proposed approach extends previous deep learning despeckling methods by applying the Conditional Variational Autoencoder (CVAE) in a generative framework. Rather than treating despeckling as a deterministic regression task from noisy to clean SAR imagery, we model it as a conditional generative process. This perspective is well aligned with the physical characteristics of SAR speckle, which introduces inherent ambiguity in recovering the underlying reflectivity. Because multiple clean images may plausibly explain a given noisy observation, despeckling can be interpreted as an ill-posed inverse problem with a distribution of valid solutions. Generative models such as the CVAE are explicitly designed to approximate such conditional distributions. By learning a latent representation of plausible despeckled outputs conditioned on noisy SAR patches, the model captures both structural priors and uncertainty in the reconstruction process.

The generative formulation enables two key advantages: the ability to produce diverse yet physically plausible reconstructions, and principled estimation of pixel-wise uncertainty through the learned latent distribution. These properties are particularly valuable in downstream flood mapping, where uncertainty in surface classification may propagate from despeckling errors. Using a VAE-based architecture also has the advantage of less-compute intensive training and inference over DDPMs, and better training stability over GANs.

Architecturally, the CVAE consists of an encoder, a latent space, and a decoder. The encoder maps the input data $y$ conditioned on $x$ to a distribution over latent variables, parameterized as a Gaussian with mean $\mu_z$ and standard deviation $\sigma_z$. A latent vector $z$ is sampled from this distribution using the reparameterization trick: $z = \mu_z + \sigma_z \odot \epsilon$, where $\epsilon \sim \mathcal{N}(0, I)$. The decoder then reconstructs the input from the latent $z$ with conditioning from $x$, producing the output $\hat{y}$. The CVAE is trained by minimizing a loss that balances reconstruction and the divergence between the learned latent distribution and the prior. The objective function is:

\begin{multline}
\mathcal{L}
= \mathbb{E}_{q_\phi(z \mid x, y)} \bigl[ \log p_\theta(y \mid z, x) \bigr] \\
- \beta \, \mathrm{KL}\!\left(
q_\phi(z \mid x, y)
\,\middle\|\, 
p_\theta(z \mid x)
\right)
\end{multline}

where the first term is the reconstruction loss and the second term is the Kullback–Leibler (KL) divergence between the approximate posterior $q_\phi(z \mid x, y)$ and the prior $p_\theta(z \mid x)$ (typically a standard normal distribution). The $\beta$ constant adjusts the tradeoff between reconstruction fidelity and the strength of regularization from the KL divergence term. The conditioning input $x$ represents the noisy SAR patch, and the input $y$ represents the multitemporal composite SAR patch. Once CVAE training is complete, conditional generation is conducted by sampling $z \sim \mathcal{N}(0, I)$ from the Gaussian prior and passing it to the decoder with $x$.

The implementation details of the CVAE are as follows. The encoder uses filter sizes $[32, 64, 128, 256, 512]$ with strided 3×3 convolutions (stride=2) for progressive downsampling, batch normalization, and LeakyReLU activations. The encoder maps the concatenated speckled input $x$ and target composite $y$ to a 200-dimensional latent space via fully connected layers for $\mu_z$ and $\log \sigma_z^2$. In the decoder, the sampled latent $z$ is projected and spatially upsampled through transposed convolutions back to the original resolution and concatenated channel-wise with the conditioning input $x$. The combined result is passed into a UNet with feature sizes $[64, 128, 256, 512]$ that outputs the reconstructed $\hat{y}$.

\subsection{Training Curriculum}
\label{section:training}
This section describes the training procedures for the multispectral and SAR water segmentation models.

Training was performed with the Adam optimizer, with learning rate and scheduler as tuned hyperparameters. Early stopping was employed with a patience of $30$ for regularization, up to a maximum of $400$ epochs. The model was checkpointed at the epoch with the lowest validation loss, and the corresponding weights were used for model evaluation. The multispectral model was trained with a single-GPU with a fixed batch size of $2048$ on an NVIDIA A100. For the SAR model, distributed multi-GPU training with 4-8 NVIDIA A100s was done for higher throughput with an effective batch size of $4096$. Input preprocessing was done on a per channel basis. The S2 band channels were kept in unnormalized reflectance values in the range $[0, 1]$ to preserve ratio information. The S2 derived water indices were also kept unnormalized. The supplementary channels that included slope and DEM were standardized using mean std normalization, except for the binary channels which were kept binary. The SAR channels were mean std normalized. For data augmentation, training patches were randomly flipped horizontally and vertically and randomly rotated $90$ to $270$ degrees to ensure the model learned generalized representations of flood boundaries regardless of specific orientations.

For the loss objective, four losses were searched during tuning: binary cross entropy (BCE) loss, combined BCE-Dice loss, Tversky loss, and Focal Tversky loss. A large problem for training was the high negative to positive class ratio, with a $6:1$ and $10:1$ ratio of negative to positive pixels in the multispectral and SAR datasets respectively. The Tversky loss was chosen to address the class imbalanced dataset, with $\alpha$ adjusting the weighting of penalties for false positives, $\beta$ for false negatives, and the constraint that $\alpha+\beta=1$. The Tversky loss is defined with respect to the Tversky Index, with the $\epsilon$ term added for stability:

\begin{multline}
\text{TI}(y, \hat{y})
= \\
\frac{
\sum_{i=1}^{N} y_i \hat{y}_i + \epsilon
}{
\sum_{i=1}^{N} \bigl(
y_i \hat{y}_i
+ \alpha (1 - y_i)\hat{y}_i
+ \beta y_i (1 - \hat{y}_i)
\bigr)
+ \epsilon
}
\end{multline}

\begin{equation}
\mathcal{L}_{\text{Tversky}} = 1 - \text{TI}(y, \hat{y})
\end{equation}

Studies have found that higher betas can lead to better generalization and results with class imbalance, with a common recommended default of $\beta=0.7$ (corresponding to $\alpha=0.3$) \cite{tverskyloss}. This decreases false negatives which are tolerated less than false positives in emergency flood mapping scenarios.

The Focal Tversky is a modification of the Tversky, defined as:

\begin{equation}
\mathcal{L}_{\text{Tversky}} = (1 - \text{TI}(y, \hat{y}))^{\frac{1}{\gamma}}
\end{equation}

where the hyperparameter $\gamma$ ranges between $[1, 3]$. In addition to similar benefits toward the class imbalance of Tversky, the loss gives increased weighting on hard examples with $\gamma > 1$, helping the model to learn to segment smaller regions that might not otherwise contribute to the loss \cite{focaltversky}. In the case of the flood dataset, it can prioritize ambiguous pixels over the dominant large water body pixels. The commonly recommended default for this purpose is $\gamma=\frac{4}{3}$.

To reduce the search space and conserve resources, the loss hyperparameters were fixed to reasonable defaults. For BCELoss and BCEDiceLoss, positive weighting was chosen to account for the class imbalance, and the negative to positive ratio of pixels in the training dataset ($6$ for multispectral, $10$ for SAR) was used for the weight value. For TverskyLoss, a $\beta$ value of $0.7$ was used based on literature recommendation. Similarly for the Focal Tversky, a $\gamma$ value of $\frac{4}{3}$ was used according to the literature.

The tuning methodology used a Centralized Bayesian Optimization algorithm (CBO) to sequentially explore the hyperparameter search space as shown in Table \ref{tab:s2_hyperparams}. The DeepHyper Python package (v0.10.0) was used for CBO with a Random-Forest surrogate model \cite{deephyper_software}. Tuning was done independently for each of 3 separate channel configurations for the S2 dataset: reflectance only, reflectance with water indices, all channels (except DEM). For the S1 SAR dataset, 2 channel configurations were tuned independently: SAR VV+VH only, and all channels (except DEM). Notably, the DEM was excluded in the all channels configuration due to overfitting from lack of diverse elevations.

For each channel configuration, the search was run until the tuning objective no longer improved, with a search-level early stopping patience of $30$. The exception was the tuning for the CVAE-despeckled SAR input models, which for fair comparison purposes, used the same tuning budget as their speckled SAR input counterparts in the UNet and UNet++ configurations.  They were tuned using $40$ and $45$ runs, respectively. The final validation F1 score with a threshold of $0.5$ was used as the tuning objective for S2 models, and the final validation AUPRC score was used as the tuning objective for S1 models.

\subsection{Despeckler Training}

This section describes the training procedure for an optional CVAE SAR despeckler that can be integrated into the SAR water segmentation pipeline.

Training occurred in two phases: pre-training, and fine-tuning. In each phase, training was performed with the Adam optimizer, with a tuned learning rate and no scheduler. In the pre-training phase, the model was trained on a large corpus of SAR images across the CONUS, with early stopping employed with a patience of $30$, up to a maximum of $400$ epochs. In the fine-tuning phase, the model was trained on a smaller subset of SAR images from regions in the S1 dataset, with early stopping patience of $80$, up to $160$ epochs. The model was checkpointed at the epoch with the lowest validation loss (the ELBO loss), with weights used for evaluation. The model was trained on 8 NVIDIA A100 GPUs with $\texttt{bfloat16}$ mixed precision with an effective batch size of $16384$. For input preprocessing, the noisy VV and VH channels were mean and std normalized. The composite VV and VH channels were mean and std normalized using the noisy VV and VH statistics. Using the same statistics ensured both noisy inputs and clean targets were represented in the same normalized space. The CVAE would learn the mapping from the noisy distribution (more variance from speckle) to the clean distribution (lower variance from averaging). For data augmentation, training patches were randomly flipped horizontally and vertically, with random rotation applied from 90 to 270 degrees.

The losses explored during tuning included: L1 loss, L2 loss, and Pseudo-Huber loss. Losses are computed per pixel, summed over all pixels within each sample, and then averaged across the batch. The Pseudo-Huber loss is a smooth approximation of the Huber loss function, defined as:

\begin{equation}
\mathcal{L}_{\delta}(r) =\delta^2 \left(\sqrt{1+\left(\frac{r}{\delta}\right)^2} - 1 \right)
\end{equation}

where $r$ is the residual of a pixel (predicted minus ground truth) and $\delta > 0$ controls the transition between quadratic and linear behavior. Pseudo-Huber has benefits in reconstruction tasks as it combines the robustness of L1 loss with the smooth optimization behavior of L2 loss. It exhibits quadratic curvature for small residuals and approximately linear growth for large residuals, resulting in bounded gradient magnitudes. This property is suitable for speckle contaminated multitemporal targets, as it prevents outliers from disproportionately influencing the loss objective. An additional adversarial loss was also explored due to blurriness early on in training. The adversarial loss was added on top of the reconstruction loss with the aim to improve the perceptual sharpness of the generated speckle free image. In the GAN-style setting, the discriminator is trained to distinguish between generated despeckled images and target multitemporal images.

Due to resource and time constraints, CVAE tuning was conducted by evaluating a small set of candidate configurations. This included different learning rates, KL annealing, loss types, and CVAE latent dimension sizes. The best set of hyperparameters was a learning rate of $0.002$, Pseudo-Huber loss with $\delta=0.3$, no KL annealing and a latent dimension size of $200$. The adversarial loss was not used since the generator learned to produce fake speckle to match the multitemporal targets rather than improve image sharpness.

Lastly, measures were taken to avoid posterior collapse, a common failure mode in VAEs. Posterior collapse occurs when the KL term vanishes and the latent variables become uninformative. To guard against this behavior, a $\beta$-annealing schedule (cyclically increasing $\beta$ from $0$ to $1$) \cite{cyclicalannealing} was included as a tunable hyperparameter. Several standard diagnostics were also monitored during training, including median KL divergence, the fraction of samples with near-zero KL, ablation-$\Delta$, and Active Units (AU). The ablation-$\Delta$ metric computes $L(z=0) - L(z=\mu)$, the difference in loss between passing $z=0$ and the encoder $z=\mu$ to the decoder. A near zero ablation-$\Delta$ indicates $z$ is ignored. The AU metric counts the number of latent dimensions with $\mu$ variance exceeding a threshold. A low AU count signals that most latent dimensions have collapsed to point masses.

\subsection{Shift-Invariant Loss}
\label{section:shift-invariant}

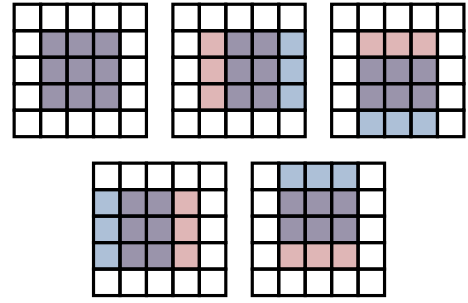
\begin{figure}[!t]
    \centering
    \begin{tikzpicture}[scale=0.35]

\definecolor{encoderBlue}{RGB}{52, 98, 155}   
\definecolor{latentTeal}{RGB}{40, 140, 135}   
\definecolor{accentRed}{RGB}{176, 74, 74}     

\tikzset{
    gridcell/.style={draw=black, line width=1.2pt},
    redcell/.style={draw=accentRed, line width=0.9pt, fill=accentRed, fill opacity=0.4},
    bluecell/.style={draw=encoderBlue, line width=0.9pt, fill=encoderBlue, fill opacity=0.4}
}

\foreach \i in {0,1,2} {
    \foreach \x in {1,2,3} {
        \foreach \y in {1,2,3} {
            \filldraw[redcell] (\i*6+\x, \y) rectangle ++(1,1);
        }
    }

    \ifnum\i=0
        \foreach \x in {1,2,3} {
            \foreach \y in {1,2,3} {
                \filldraw[bluecell] (\i*6+\x, \y) rectangle ++(1,1);
            }
        }
    \else\ifnum\i=1
        \foreach \x in {2,3,4} {
            \foreach \y in {1,2,3} {
                \filldraw[bluecell] (\i*6+\x, \y) rectangle ++(1,1);
            }
        }
    \else\ifnum\i=2
        \foreach \x in {1,2,3} {
            \foreach \y in {0,1,2} {
                \filldraw[bluecell] (\i*6+\x, \y) rectangle ++(1,1);
            }
        }
    \fi\fi\fi

    \foreach \x in {0,1,2,3,4} {
        \foreach \y in {0,1,2,3,4} {
            \draw[gridcell] (\i*6+\x, \y) rectangle ++(1,1);
        }
    }
}

\begin{scope}[yshift=-6cm,xshift=3cm]
    \foreach \i in {0,1} {
        \foreach \x in {1,2,3} {
            \foreach \y in {1,2,3} {
                \filldraw[redcell] (\i*6+\x, \y) rectangle ++(1,1);
            }
        }

        \ifnum\i=0
            \foreach \x in {0,1,2} {
                \foreach \y in {1,2,3} {
                    \filldraw[bluecell] (\i*6+\x, \y) rectangle ++(1,1);
                }
            }
        \else\ifnum\i=1
            \foreach \x in {1,2,3} {
                \foreach \y in {2,3,4} {
                    \filldraw[bluecell] (\i*6+\x, \y) rectangle ++(1,1);
                }
            }
        \fi\fi

        \foreach \x in {0,1,2,3,4} {
            \foreach \y in {0,1,2,3,4} {
                \draw[gridcell] (\i*6+\x, \y) rectangle ++(1,1);
            }
        }
    }
\end{scope}

\end{tikzpicture}
    \caption{Shift-invariant loss. Red indicates the prediction from S1; Blue indicates the ground truth from S2. The shape of each window is reduced to $3 \times 3$ within a $5 \times 5$ patch for illustration purposes only.}
    \label{fig:shift-invariant}
\end{figure}

Geolocation uncertainty introduces residual spatial misregistration between multispectral and SAR imagery in the SAR dataset. Reported geolocation uncertainty is approximately 20m at $2\sigma$ and 7m at $3\sigma$ for S2 L2A and S1 IW GRD products respectively \cite{sentinel2, s1productdef}. Consequently, shifts of up to \(20 \, \text{m}\) in landmarks (two pixels at 10m resolution) were qualitatively observed during dataset inspection. Since the SAR model is trained using labels derived from temporally coincident S2 captures, it is important to prevent the model from learning systematic spatial bias induced by the misregistration errors.

The shift-invariant loss addresses this residual misregistration by introducing an alignment-aware training objective that tolerates spatial translations between prediction and label. The loss allows the predicted S1 water mask to be shifted both vertically and horizontally to match the S2 ground truth, as shown in Figure~\ref{fig:shift-invariant}.

Let $S_{dx,dy}(Y) \in \mathbb{R}^{H \times W}$ be the windowing operator that selects the $H \times W$ subregion of $Y$ that is offset by $(dx, dy)$ from the centered window, where $dx, dy \in \{-d, \ldots, d\}$. The shift-invariant loss \( L'(Y, \hat{Y}) \) is defined as:

\begin{multline}
L'(Y, \hat{Y})
= \\
\min_{dx,dy \in \{-d,\ldots,d\}}
\sum_{i=1}^{H} \sum_{j=1}^{W}
L\!\left(
S_{dx,dy}(Y)_{ij},\,
\hat{Y}_{ij}
\right)
\label{eq:loss-shift}
\end{multline}

where \( Y \) represents the ground truth label, \( \hat{Y} \) is the prediction, \( d \) is the shift tolerance, and $L(\cdot)$ denotes any pixel-wise segmentation loss. Out of $(2d + 1) \times (2d+1)$ possible windows within the label, the window yielding the lowest loss is used for backpropagation. The shift-invariant objective thus composes a discrete alignment operator with any differentiable segmentation loss.

For training the SAR model, each of the loss functions listed in \ref{section:training} above was adapted with shift invariance to handle spatial misalignment between SAR imagery and S2-derived labels. To accommodate upwards of a 20m shift vertically or horizontally in either direction, a shift tolerance of $d=2$ pixels was applied to the central $64 \times 64$ window of $68 \times 68$ patches in the SAR dataset.

\subsection{Evaluation}

For model evaluation on the S2 multispectral and S1 SAR dataset, pixel-level precision, recall, F1 score, Intersection-over-Union (IoU), and Area Under the Precision-Recall Curve (AUPRC) were reported. Due to extreme class imbalance between wet and dry pixels, accuracy is not a meaningful indicator of model performance and hence excluded. AUPRC is used as the primary evaluation metric because it directly characterizes the precision-recall trade-off under severe class imbalance and is invariant to choice of decision thresholds. IoU and F1 score are reported at a threshold of $0.5$ to facilitate comparison with prior flood-mapping literature. Operational F1 scores from validation calibrated thresholds are additionally reported. Metrics are reported overall and stratified by NLCD land cover and SCL scene classes to assess performance variation across land cover types and imaging conditions.

For SAR despeckling, reconstruction quality is evaluated using Peak Signal-to-Noise Ratio (PSNR), Structural Similarity Index Measure (SSIM), and Equivalent Number of Looks (ENL). PSNR measures pixel-wise fidelity to the reference image, while SSIM evaluates structural and perceptual similarity. In contrast, ENL is a no-reference metric that quantifies speckle suppression within homogeneous regions. It is defined as the ratio of the squared mean to the variance of pixel intensities, with higher values indicating improved speckle suppression. PSNR and SSIM are computed against multitemporal composite references, which are not true speckle-free ground truth but serve as a practical surrogate. ENL is separately computed on 75 manually selected homogeneous patches from test tiles, following standard SAR despeckling evaluation practice.

For all results reported in section \ref{section:results}, a one tailed t-test is performed with $\alpha = 0.05$ when asserting statistical significance of one metric over another.

\subsection{Software and Implementation Details}

\noindent\textbf{Geospatial data processing}
\begin{itemize}
  \item \textbf{Python} (v3.11): data download and preprocessing.
  \item \textbf{Rasterio} (v1.3.10): reprojection, resampling, and raster I/O.
  \item \textbf{GIMP}: raster editing; generation of water-presence masks.
  \item \textbf{QGIS}: visualization and comparison with high-resolution imagery.
\end{itemize}

\noindent\textbf{Machine-learning experiments}
\begin{itemize}
  \item \textbf{PyTorch} (v2.3.0): model implementation and training.
  \item \textbf{NumPy} (v1.26.4): numerical computing and array operations.
  \item \textbf{Matplotlib} (v3.8.4): plots and training curves.
  \item \textbf{DeepHyper} (v0.10.0): Bayesian optimization for hyperparameter tuning.
  \item \textbf{Weights \& Biases} (v0.17.1): experiment tracking.
\end{itemize}

\section{Results}
\label{section:results}

\begin{figure*}[!t]
    \centering
    \includegraphics[width=15cm]{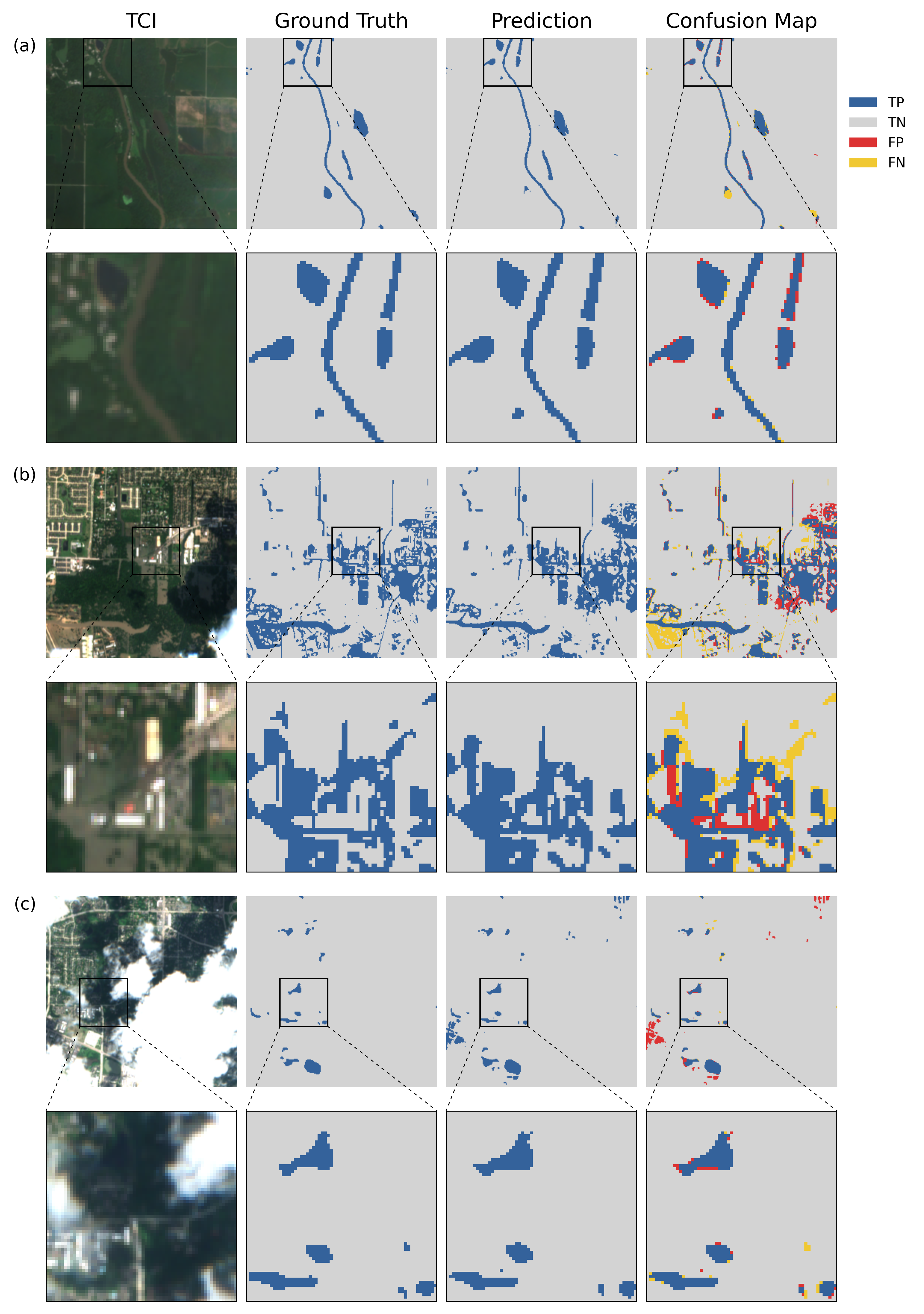}
    \caption{Prediction of UNet++ trained on all channels with $t = 0.75$ on S2 4km x 4km tiles from test set. Example tiles were chosen to highlight performance in (a) haze, (b) urban areas, and (c) cloud shadow. The zoomed in area is $64 \times 64$ pixels. Confusion matrix categories (TP = True Positive, TN = True Negative, FP = False Positive, FN = False Negative) are shown in the column labeled Confusion Map.}
    \label{fig:s2hardexamples}
\end{figure*}

\begin{table*}[!t]
    \centering
    \caption{Benchmarks for S2 multispectral models. Standard error is reported for mean AUPRC across $N=10$ trials.}
    \label{tab:s2_benchmarks}
    \begin{tabular}{l c c c c}
    \toprule
    \textbf{Model} & \textbf{Channels} & \textbf{AUPRC} & \textbf{F1} ($t = 0.5$) & \textbf{IoU} ($t = 0.5$) \\
    \midrule
    UNet++ & All & $\mathbf{0.9557} \pm \mathbf{0.0017}$ & 0.8746 & 0.7773 \\
    UNet++ & S2 + Indices & $0.9540 \pm 0.0028$ & 0.8675 & 0.7666 \\
    UNet & S2 + Indices & $0.9489 \pm 0.0026$ & 0.8549 & 0.7469 \\
    UNet & S2 Only & $0.9445 \pm 0.0064$ & 0.8504 & 0.7401 \\
    UNet & All & $0.8762 \pm 0.0080$ & $\mathbf{0.8807}$ & $\mathbf{0.7871}$ \\
    UNet++ & S2 Only & $0.8121 \pm 0.0255$ & 0.8082 & 0.6801 \\
    \bottomrule
    \end{tabular}
\end{table*}

The test-set performance on the S2 dataset is summarized in Table \ref{tab:s2_benchmarks}. By AUPRC, the best-performing S2 model was UNet++ trained with all channels except DEM. When evaluated without calibration using a fixed threshold of $0.5$, the UNet trained with all channels achieved the highest F1 and IoU scores. Across architectures, the UNet++ model achieved statistically significant improvements ($p < 0.05$) in AUPRC scores over the UNet model in the input-rich channel configuration that contained additional water indices and supplementary data. Incorporating water indices also led to systematic improvements in AUPRC relative to using reflectance channels alone.

\begin{sidewaysfigure*}[!t]
    \centering
    \includegraphics[width=\textheight]{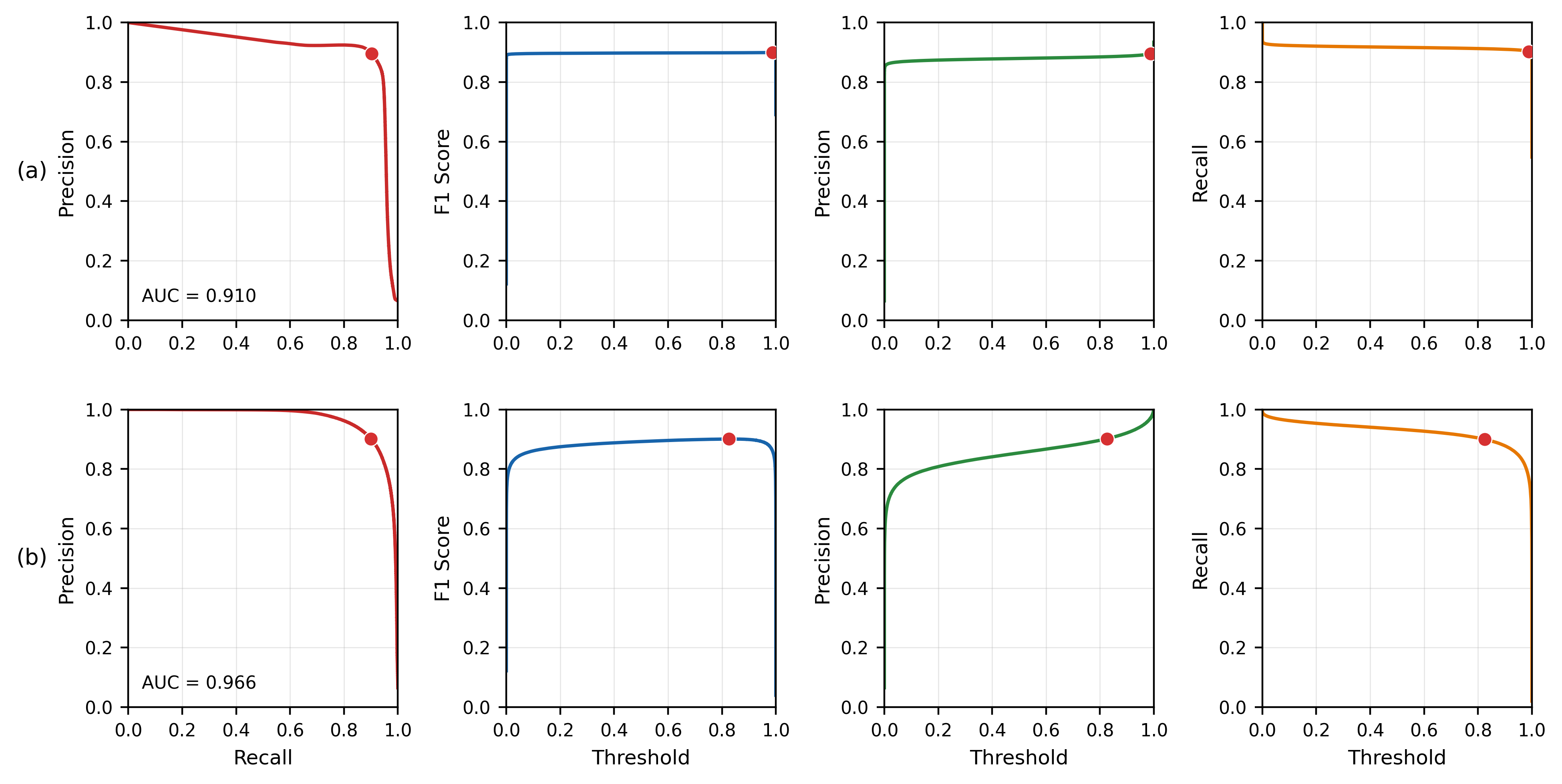}
    \caption{Each row shows in order the Precision-Recall curve, F1 score, precision, and recall plots on the test set using all channels for (a) UNet, (b) UNet++. The red dot indicates the threshold with the max F1 score.}
    \label{fig:s2prcurves}
\end{sidewaysfigure*}

Precision-recall curves in Figure \ref{fig:s2prcurves} further highlight differences in architecture: UNet++ with all channels exhibited a broader curve and a more favorable precision-recall tradeoff than the corresponding UNet model. Furthermore, after F1-optimal threshold calibration on the validation set, UNet++ achieved a higher operational F1 score on the test set. The optimal validation-set threshold for UNet++ was $0.75$, resulting in a test-set (operational) F1, recall, and precision of $0.8998$, $0.9112$, and $0.8886$ respectively, compared to a calibrated threshold of $0.20$ and an operational F1, recall, and precision of $0.8967$, $0.9207$, and $0.8740$ respectively, for UNet (Table \ref{tab:calibrated_threshold}).

The models demonstrate robustness under challenging land classes and weather conditions. Quantitatively, the top performing S2 model achieves F1 scores of $0.877$ for haze, $0.781-0.877$ for shadow, and $0.787$ for urban structure (Table \ref{tab:class_f1}). Qualitatively, the calibrated UNet++ predictions further illustrate this robustness, accurately delineating thin streams and 1-2 pixel-wide water bodies even within haze, shadowed regions, and dense urban scenes (Figure \ref{fig:s2hardexamples}).

However, issues persist in generalizing to large datasets. A qualitative evaluation of predictions across 10,000+ unseen tiles during weak labeling of the S1 dataset (using the S2 UNet on all channels with threshold $t = 0.5$ and test F1, recall, and precision of $0.8977$, $0.9169$, and $0.8794$) identified prediction issues stemming from solar-illumination conditions and water confounders (e.g., shadow, asphalt, and wet sand). Specifically, false negatives came from water bodies exhibiting specular reflection in several tiles. Consistent false positives resulted from mountainous regions or coastal scenes, where shadow pixels in and around dense forests, or pixels from wet sand, were commonly mislabeled as water. Urban structures such as roads and pavements also contributed to false positives, although inconsistently.

\begin{table*}[t]
    \centering
    \caption{Benchmarks for S1 SAR models trained with shift-invariant loss and regular loss. Standard error is reported for mean AUPRC across $N=10$ trials. F1 and IoU are computed using a fixed threshold of $t = 0.5$. Metrics are computed using the label window (i.e. shift) that minimizes model loss, not necessarily the central window.}
    \label{tab:s1_benchmarks}
    \begin{tabular}{l c c c c c c c}
    \toprule
    \textbf{Model} & \textbf{Channels} 
    & \multicolumn{3}{c}{\textbf{Shift-Invariant Loss}}
    & \multicolumn{3}{c}{\textbf{Regular Loss}} \\
    \cmidrule(lr){3-5} \cmidrule(lr){6-8}
    & 
    & \textbf{AUPRC} 
    & \textbf{F1}
    & \textbf{IoU}
    & \textbf{AUPRC} 
    & \textbf{F1}
    & \textbf{IoU} \\
    \midrule
    UNet++ & All 
    & $\mathbf{0.8397} \pm \mathbf{0.0034}$ & 0.7344 & 0.5804
    & $\mathbf{0.8426} \pm \mathbf{0.0030}$ & 0.7611 & 0.6144 \\
    UNet & All 
    & $0.8369 \pm 0.0019$ & $\mathbf{0.7378}$ & $\mathbf{0.5847}$
    & $0.8414 \pm 0.0025$ & $\mathbf{0.7635}$ & $\mathbf{0.6175}$ \\
    UNet++ & VV+VH 
    & $0.7797 \pm 0.0017$ & 0.6950 & 0.5327
    & $0.7815 \pm 0.0016$ & 0.7324 & 0.5779 \\
    UNet & VV+VH 
    & $0.7795 \pm 0.0020$ & 0.6998 & 0.5383
    & $0.7808 \pm 0.0012$ & 0.6844 & 0.5202 \\
    \bottomrule
    \end{tabular}
\end{table*}

Table \ref{tab:s1_benchmarks} summarizes the test-set performance of SAR models trained with shift invariant loss and regular loss on the  S1 dataset. Metrics in the table are calculated using labels selected by shift invariant loss. By AUPRC, the best performing SAR model was the UNet++ with all channels except DEM. Without calibration, the UNet with all channels yielded the highest F1 and IoU score with a fixed threshold of $0.5$. In contrast to the S2 dataset, UNet++ did not show statistically significant differences ($p > 0.05$) to UNet on the S1 dataset in any channel configuration. However, the all channels configuration showed statistically significant improvements ($p < 0.05$) in AUPRC over the VV+VH only configuration. The optimal validation-set threshold for the UNet++ on all channels was $0.85$, resulting in a test-set (operational) F1, recall, and precision of $0.7532$, $0.6361$, and $0.9235$ respectively (Table \ref{tab:calibrated_threshold}).

\begin{figure}[!htp]
    \centering
    \includegraphics[width=0.9\columnwidth]{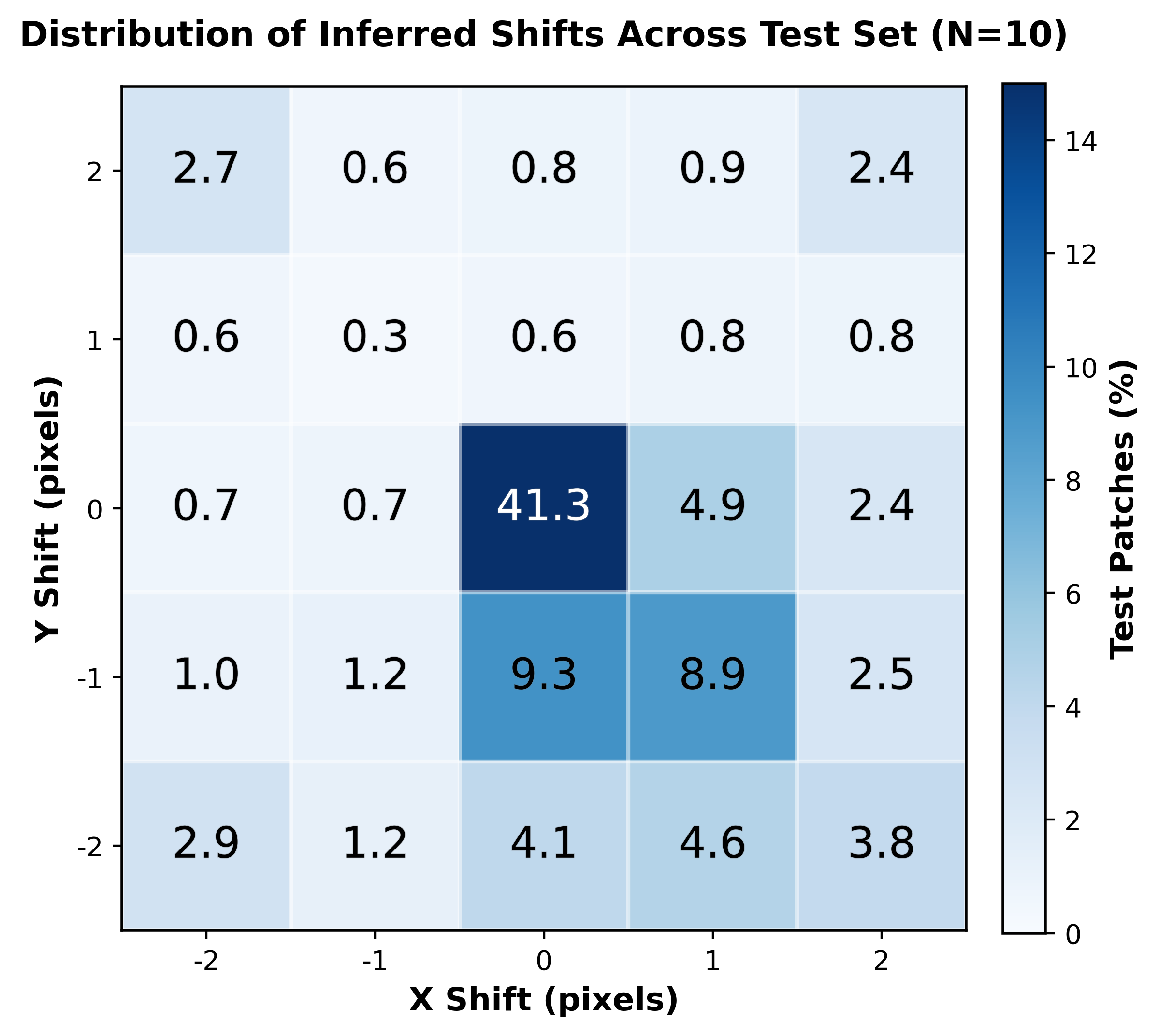}
    \caption{Percentage distribution of inferred $(x,y)$ pixel shifts of the ground-truth label (S2) relative to SAR (S1), reported as the mean across $N=10$ trials for the best performing UNet++ trained with shift-invariant loss. For each test sample, the shift is estimated by selecting the $(x,y)$ offset that minimizes the prediction-label error.}
    \label{fig:shiftdist}
\end{figure}

Importantly, shift-invariant loss is not intended to improve performance under evaluation regimes that retain image to label misregistration (i.e. using uncorrected labels) or that fail to penalize misaligned predictions (i.e. using shift invariant loss aligned labels). Rather, it is designed to prevent the model from learning spurious spatial biases induced by misaligned training labels. Consistent with this design, models trained with shift-invariant loss do not necessarily achieve higher AUPRC than those with regular loss when evaluated against uncorrected or shift invariant loss aligned labels (in fact, there is no statistically significant difference in a two-tailed t-test), but instead quantitatively and qualitatively produce predictions that better respect coherent SAR backscatter structure. A visual evaluation of inference on SAR tiles showed that labels tended to be shifted down and to the right, and that shift-invariant loss models corrected for the shift by making a similar prediction up and to the left. This was confirmed by the distribution of shifts across the test dataset shown in Figure \ref{fig:shiftdist}, with a bias toward 1-2 pixel shifts in the south and east directions. A quantitative analysis of how this behavior translates into improved performance once alignment error is removed is established through a controlled ablation study in section \ref{section:ablation}.

\begin{table}[!t]
    \centering
    \caption{Benchmarks for S1 SAR despeckling across VV and VH polarizations.}
    \label{tab:despeckle_benchmarks}
    \setlength{\tabcolsep}{4pt}
    \begin{tabular}{l c c c c c c}
    \toprule
    & \multicolumn{3}{c}{\textbf{VV}} & \multicolumn{3}{c}{\textbf{VH}} \\
    \cmidrule(lr){2-4} \cmidrule(lr){5-7}
    \textbf{Despeckling}
    & \textbf{PSNR}
    & \textbf{SSIM}
    & \textbf{ENL}
    & \textbf{PSNR}
    & \textbf{SSIM}
    & \textbf{ENL} \\
    \midrule
    CVAE 
    & $\mathbf{25.92}$ & $\mathbf{0.63}$ & $\mathbf{188.7}$
    & $\mathbf{27.09}$ & $\mathbf{0.66}$ & $\mathbf{379.5}$ \\
    Enhanced Lee 
    & 24.65 & 0.57 & 26.6
    & 25.81 & 0.60 & 32.1 \\
    Raw 
    & 22.03 & 0.42 & 6.1
    & 22.92 & 0.42 & 6.6 \\
    \bottomrule
    \end{tabular}
\end{table}

\begin{figure*}[t]
    \centering
    \includegraphics[width=0.7\textwidth]{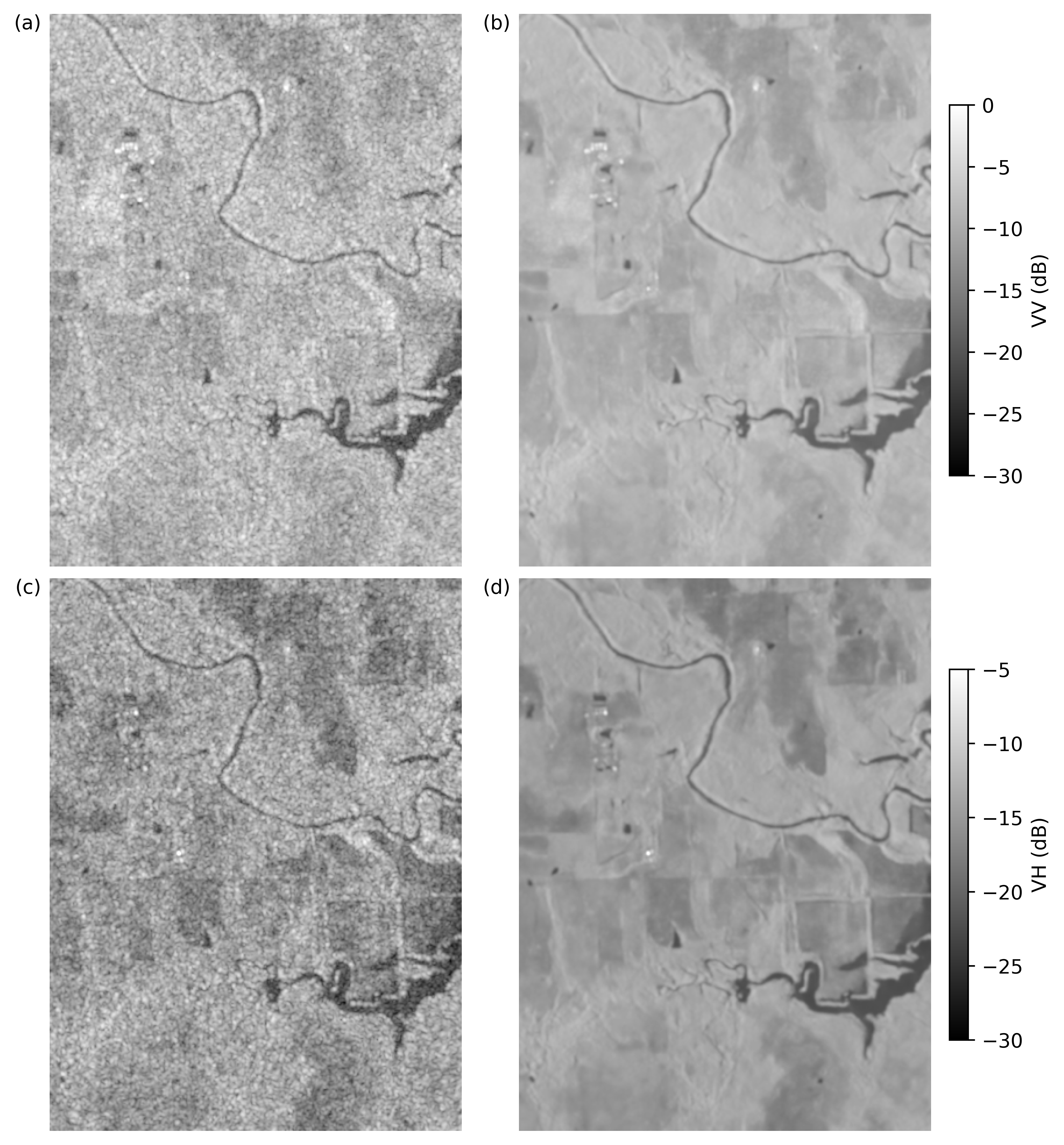}
    \caption{Example S1 SAR 4km x 4km tile from test set. The VV and VH polarizations are shown before and after CVAE despeckling. (a) VV (noisy). (b) VV (despeckled with the trained CVAE). (c) VH (noisy). (d) VH (despeckled with the trained CVAE).}
    \label{fig:cvaesar}
\end{figure*}

The results for the CVAE despeckler on the S1 multitemporal fine-tuning dataset are shown in Table \ref{tab:despeckle_benchmarks}. The CVAE outperforms the Enhanced Lee filter in PSNR, SSIM, and ENL despeckling metrics. Furthermore, qualitative evaluation of CVAE despeckled images on the test set showed a high level of perceptual sharpness relative to the Enhanced Lee filter, preserving fine structures and edges in noisy SAR images while smoothing homogeneous regions. Bright scatterers were also preserved. Examples of CVAE despeckled VV and VH images can be seen in Figure \ref{fig:cvaesar}.

While these results demonstrate that the CVAE substantially improves SAR image quality, improvements in despeckling do not necessarily correspond to gains in flood mapping performance. To assess the effect of CVAE-based despeckling on downstream flood mapping, a controlled CVAE ablation study is conducted in section \ref{section:ablation}.

\subsection{Ablation Study}
\label{section:ablation}

\begin{table*}[t]
    \centering
    \caption{Benchmarks for S1 SAR models tuned independently on shift invariant loss and regular loss modes. Standard error is reported for mean AUPRC across $N=10$ trials. Evaluation is performed on S2 flood labels manually aligned with multiple GCPs. All channels are used.}
    \label{tab:s1_benchmarks_ablation}
    \begin{tabular}{l c c c c}
    \toprule
    \textbf{Model} & \textbf{Shift Invariant Loss} & \textbf{AUPRC} & \textbf{F1} ($t = 0.5$) & \textbf{IoU} ($t = 0.5$) \\
    \midrule
    UNet++ & \textbf{True} & $\mathbf{0.7790} \pm \mathbf{0.0049}$ & $\mathbf{0.6976}$ & $\mathbf{0.5357}$ \\
    UNet & \textbf{True} & $\mathbf{0.7782} \pm \mathbf{0.0025}$ & $\mathbf{0.7005}$ & $\mathbf{0.5390}$ \\
    UNet++ & False & $0.7651 \pm 0.0026$ & 0.6893 & 0.5259 \\
    UNet & False & $0.7641 \pm 0.0029$ & 0.6845 & 0.5203 \\
    \bottomrule
    \end{tabular}
\end{table*}

An ablation study was performed on the shift invariant loss training method to identify whether it improves performance on SAR aligned labels. 8 diverse tiles with flooding were selected from the test set, and the weak labels for each were manually aligned with the corresponding SAR image using 5-6 Ground Control Points (GCP) between the SAR and S2 images. The shift distance was recorded for both x and y directions. For each direction, the median of all GCP shifts in meters was taken and rounded to the nearest number of pixels (nearest 10m). The label pixels were then aligned horizontally and vertically according to these values and preprocessed into patches. Finally, the best tuned hyperparameters using shift invariant loss for each model and the best tuned hyperparameters using regular loss for each model were evaluated on the held out set of alignment corrected patches. The benchmark results in Table \ref{tab:s1_benchmarks_ablation} show that the shift invariant loss trained models exhibited statistically significant improvements ($p < 0.05$) in AUPRC, F1 Score, and IoU in both the UNet++ and the UNet, indicating stronger alignment to SAR flood imagery.

\begin{table*}[!t]
    \centering
    \caption{Benchmarks for S1 models with despeckled SAR data. Standard error is reported for mean overall AUPRC across $N=10$ trials. Note that the metrics are calculated using the shifted label with lowest model loss, not necessarily the central window.}
    \label{tab:s1_filter_benchmarks}
    \begin{threeparttable}
    \begin{tabular}{l c c c c c}
    \toprule
    \textbf{Model} & \textbf{Despeckling} & \multicolumn{4}{c}{\textbf{AUPRC}} \\
    \cmidrule(lr){3-6}
     &  & \textbf{Overall} & \textbf{Forest} & \textbf{Cultivated} & \textbf{Herbaceous} \\
    \midrule
    UNet++ & CVAE & $\mathbf{0.8382}^{\dagger} \pm \mathbf{0.0019}$ & $\mathbf{0.4172}$ & $\mathbf{0.4881}$ & $\mathbf{0.6774}$ \\
    UNet++ & None & $0.8381 \pm 0.0019$ & 0.4138 & 0.4820 & 0.6755 \\
    UNet++ & Enhanced Lee & $0.8329 \pm 0.0023$ & 0.3895 & 0.4826 & 0.6516 \\
    \bottomrule
    \end{tabular}
    \begin{tablenotes}
        \footnotesize
        \item[$\dagger$] Significantly higher than Enhanced Lee ($p < 0.05$).
    \end{tablenotes}
    \end{threeparttable}
\end{table*}

An ablation study was also performed for the CVAE despeckler to identify effects on downstream flood mapping. For fair comparison, the UNet++ trained with CVAE despeckled SAR and Enhanced Lee despeckled SAR were allocated the same tuning budget as the UNet++ with raw SAR data, with a total of 41 runs each. While the mean overall AUPRC was highest for the CVAE despeckled SAR data, the results in Table \ref{tab:s1_filter_benchmarks} did not show a statistically significant gain ($p > 0.05$) over the model trained on the raw SAR data. However, the gain was statistically significant ($p < 0.05$) over the model trained on the Enhanced Lee filtered SAR data. The CVAE also showed statistically significant ($p < 0.05$) improvements in AUPRC in forest, cultivated, and herbaceous NLCD land classes over the Enhanced Lee filter.

\begin{figure*}[!t]
    \centering
    \includegraphics[width=15cm]{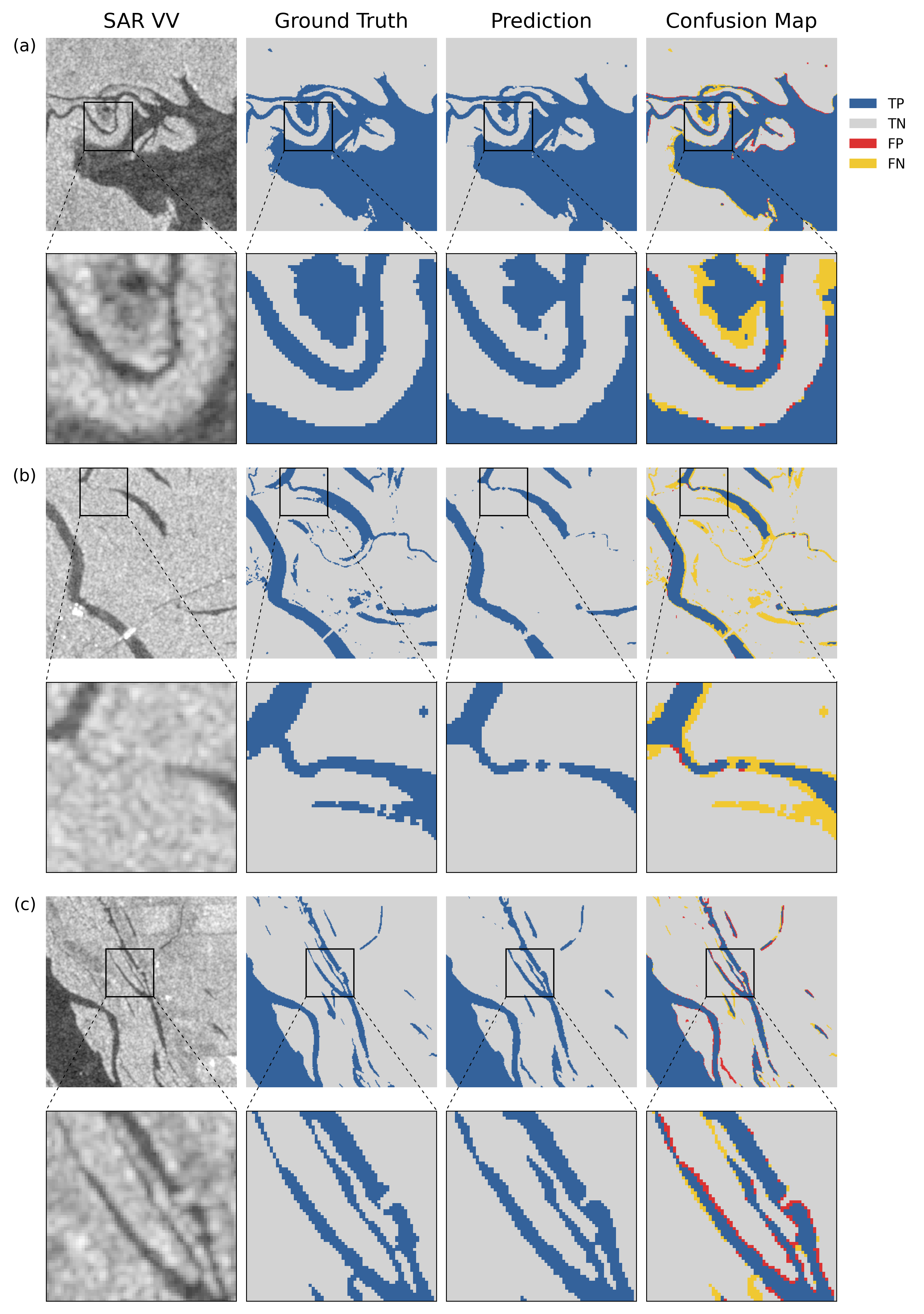}
    \caption{Prediction of UNet++ trained with shift invariant loss on all channels with $t = 0.85$ on S1 4km x 4km tiles from test set. Ground truth are manually corrected S2 labels and GCP aligned to the SAR image. The zoomed in area is $64 \times 64$ pixels.}
    \label{fig:s1examples}
\end{figure*}

\begin{figure*}[!t]
    \centering
    \includegraphics[width=15cm]{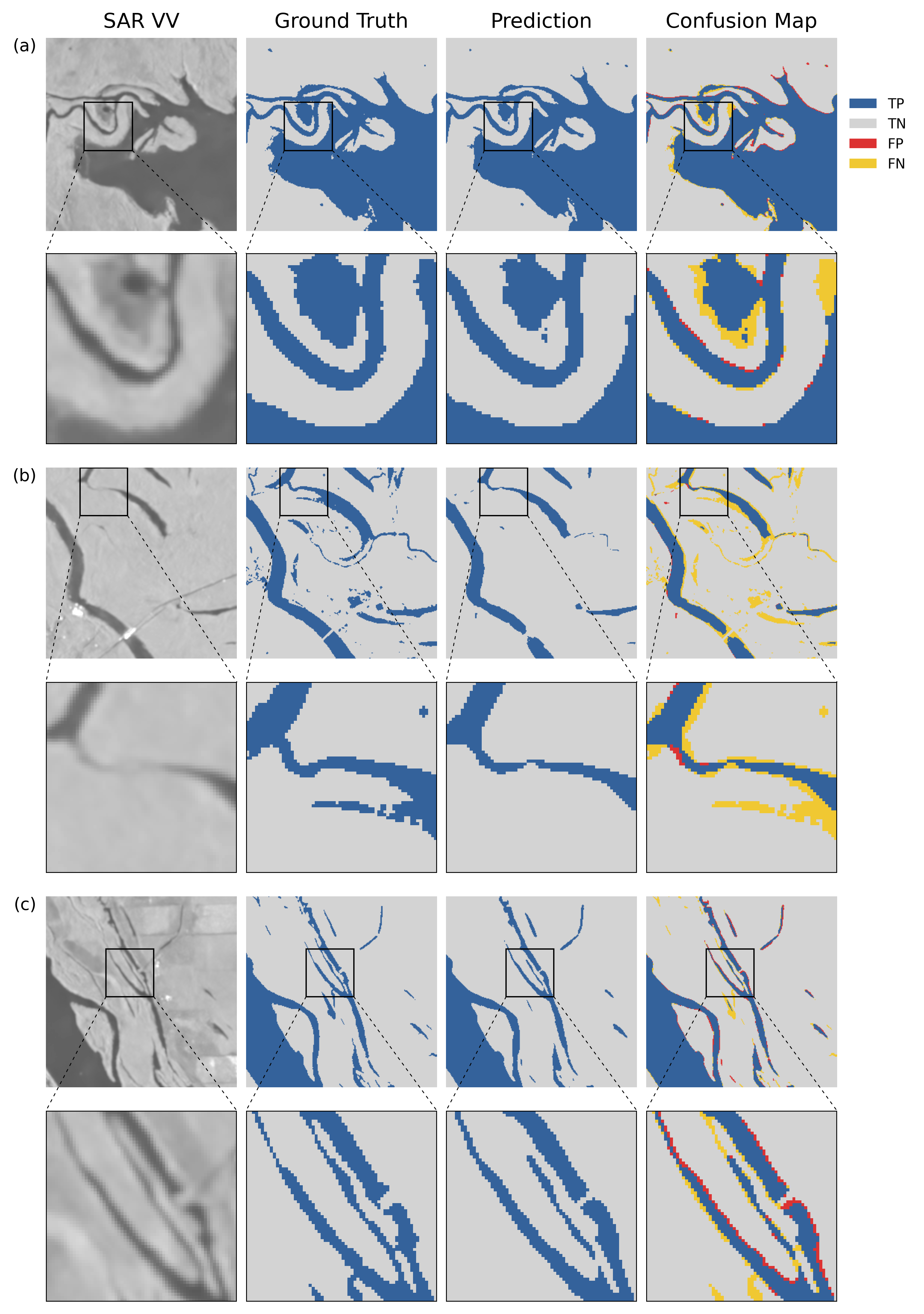}
    \caption{Prediction of UNet++ trained on CVAE despeckled images with shift invariant loss on all channels with $t = 0.88$ on S1 4km x 4km tiles from test set. Ground truth are manually corrected S2 labels and GCP aligned to the SAR image. The zoomed in area is $64 \times 64$ pixels.}
    \label{fig:s1cvaeexamples}
\end{figure*}

Qualitative evaluations of the the raw SAR model and CVAE despeckled SAR model predictions in Figure \ref{fig:s1examples} and Figure \ref{fig:s1cvaeexamples} show comparative performance in water body detection, with the exception that the CVAE SAR model produces smoother water boundaries. It is also notable that the SAR image water boundaries frequently appear interior to the S2 label water boundaries, which is discussed later as a limitation.

\section{Discussion}

\subsection{S2 and S1 Segmentation}

Segmentation performance on the S2 dataset suggests that UNet++ is better suited than UNet for exploiting diverse geospatial inputs. When all channels (including water indices and supplementary channels) were used, UNet++ yielded statistically significant increases in AUPRC and improved precision–recall characteristics. A likely explanation is that the nested and dense skip connections improve multi-scale feature fusion between the encoder and decoder, which may be beneficial for integrating heterogeneous spatial and semantic information in multispectral flood mapping.

However, despite the advantage of UNet++ on S2, UNet was selected to generate weak labels on the S1 dataset. The choice reflects differing priorities rather than a contradiction in performance. In the setting of S1 dataset labeling, recall was prioritized over precision to minimize missed flooded regions in imagery. The UNet using all channels with a fixed threshold of $t = 0.5$ was explicitly chosen as it had a more appropriate recall–precision tradeoff than the UNet++ with a calibrated operating point of $t = 0.75$ (see Table \ref{tab:calibrated_threshold}). Although alternative thresholds could potentially improve UNet++ performance in this regime, the UNet configuration offered sufficient fidelity for the purposes of this experiment and was therefore adopted.

The results also show that investing in S2 label quality can push models to attain fidelity at the 10m resolution. This is evidenced by the strong quantitative performance of the top S2 UNet++ model with a $95.6\%$ AUPRC score and $90.0\%$ calibrated F1 (Table \ref{tab:s2_benchmarks} and \ref{tab:calibrated_threshold}) on the S2 dataset, in addition to the qualitative observations of single pixel wide waterbody segmentation (for example, see Figure \ref{fig:s2hardexamples}). It is likely this fidelity in S2 cannot be achieved with traditional thresholded water masks that lack small-scale waterbodies. Many 10 meter wide waterbodies in the dataset such as ponds or streams identified through high resolution imagery would otherwise have been excluded due to low water indices in thresholded water masks.

Furthermore, including diverse challenging scenes such as haze, cloud shadow, and dense urban areas in the S2 dataset helped improve model robustness. This is indicated by the S2 model F1 score of $87.7\%$ for water prediction in haze and comparable performance in shadow and urban classes (Table \ref{tab:class_f1}). Model robustness in these regimes ensures lower label noise in the S1 dataset weak labels. Yet, there are still limitations. For heavily forested tiles, S2 images are too low in resolution and spectral separability to reliably determine water presence through canopies and shade. As a result, during manual annotation of wooded areas, only waterbodies in clearings or large diameter rivers or streams could be labeled. Hence, limited manual annotation of water in specific contexts such as dark wooded regions may inhibit the generalization of the model to those cases.

A promising future avenue for improving robustness is to incorporate broader spatial context beyond local patches (i.e., the smaller $64 \times 64$ inputs used for inference), potentially at the scale of the full S2 scene. Conditioning the model on auxiliary scene-level indicators of haze, cloud cover, or shadow may allow the model to adapt its sensitivity under varying atmospheric and illumination conditions, thereby reducing omission and commission errors.

\subsection{Shift Invariant Loss}

In contrast to the S2 results, the S1 benchmarks suggest that architecture choice may play a smaller role in SAR flood mapping performance. Across two channel configurations, no statistically significant difference was observed between UNet and UNet++ in AUPRC. This observation is consistent with prior S1 deep learning segmentation studies that have reported limited or no systematic gains from deeper architectures \cite{bereczky}. These findings suggest that improvements in SAR flood mapping may depend more strongly on training objectives and data alignment strategies than on increased architectural depth.

One of the ways to improve the SAR model is by combating the biases of S2 and S1 misalignment. This misalignment appears in the dataset as a systematic south-eastward displacement of approximately 1–2 pixels, as evidenced by quantitative analysis of the shift distribution. This directional bias is consistent with known geolocation uncertainties in S2 and S1 products. S2 data quality reports from 2024 and 2025 indicate a consistent mean Across-Track (ACT) bias of +3.56-3.66m \cite{sentinel2_l1c_dqr_2024, sentinel2_l1c_dqr_2025}, while out-of-the-box S1 IW products exhibit a mean azimuth bias of +2.8m and a negative mean range bias in geolocation uncertainty of -3.2m \cite{s1geocoding}. Given the diagonally oriented satellite ground tracks over the CONUS from NE to SW, these biases jointly induce a predominant south-eastward displacement, in agreement with the observed shift distribution.

The results show that the proposed shift-invariant loss objective helps to correct for this bias during training without a reduction in label quality. The novel shift invariant loss significantly increased trained SAR model performance on GCP-aligned labels across UNet and UNet++ architectures, while maintaining similar AUPRC scores to regular loss counterparts when shifts effects are ignored. Thus, the modification of the loss objective to allow for pixel shifts between the prediction and the ground truth improves model SAR alignment when trained on S2 labels.

A potential criticism is that the observed directional misalignment could be corrected through additional product post-processing, rather than implicitly via a shift-invariant loss. However, while more rigorous post-processing could reduce the mean offset between S1 imagery and S2-derived labels, such corrections are not guaranteed to eliminate misalignment at the tile level. In practice, residual errors post-refinement will still remain from topographic and scene-dependent variability \cite{coregistration1}, and stochastic geolocation noise \cite{sentinel2_l1c_dqr_2024, sentinel2_l1c_dqr_2025}, resulting in non-uniform shifts that cannot be corrected with a global transformation. Moreover, there are a variety of possible sources of misregistration, from resampling kernels, different processing baselines, orbit errors \cite{sargeolocation}, among others, each with their own processing challenges. As a consequence, post-processing carries a brittle dependency on the effectiveness and scope of the correction procedure that may not be able to universally eliminate the misalignment bias.

In contrast, embedding shift invariance directly into the training objective encodes the known structure of possible translations while avoiding a higher-dimensional correction problem. This formulation ensures consistent supervision across models and prevents situations in which some architectures implicitly learn the shift while others fail to do so. The shift-invariant loss therefore mitigates inductive bias in the training data from leaking into the learned model, providing a robust solution to the alignment problem.

\subsection{CVAE Despeckler}

The results demonstrate that the CVAE architecture trained on noisy and multitemporal image pairs, is a strong SAR generative despeckling model over classical filters. The CVAE significantly improved over Enhanced Lee across all despeckling metrics, while showing strong qualitative preservation of structure and detail. Compared to deterministic despeckling, it has the advantage of deterministic inference using $z = 0$ in the decoder as well as stochastic inference using $z \sim \mathcal{N}(0, I)$ to model a distribution of possible despeckled images under uncertainty.

Yet, despite high quality despeckling capabilities, a surprising finding is that adding the generative CVAE despeckler upstream of the SAR model did not lead to statistically significant gains in the AUPRC score over raw SAR input. This may be caused by sensor differences with the S2 weak labels, with SAR water boundaries qualitatively observed to be interior to the S2 label water boundaries. Even with geolocation sensor misalignments resolved, there still exists geometric distortions in the SAR image from slant range geometry. Foreshortening, layover, and radar shadow distort curvature and warp water boundaries non-uniformly. These distortions are further compounded by DEM-dependent orthorectification errors. As a result, even accurately despeckled SAR imagery may exhibit water borders that are physically consistent in SAR geometry but geometrically incompatible with nadir-view S2 labels. The label noise from geometric mismatch is especially detrimental for border-sensitive metrics such as AUPRC, effectively masking potential performance gains from despeckler-improved boundary delineation.

In order to properly evaluate the CVAE despeckling on downstream flood mapping performance, further work must be done to correct for these geometric differences. A generative technique could be explored to empirically perform coregistration of SAR and S2 images using incidence angle among other parameters in the same way shift invariant loss was employed to account for residual quality assurance and quality control uncertainty.

\section*{Author Contributions}

Conceptualization: DM, JF; Data curation: DM, SP; Formal analysis: DM; Funding acquisition: JF, EY; Investigation: DM; Methodology: DM, JF, AG; Software: DM; Supervision: JF, EY; Validation: DM; Visualization: DM, JF; Writing - original draft: DM; Writing - review \& editing: DM, JF, SP, AG, EY.

\section*{Acknowledgments}
The authors gratefully acknowledge the high-performance computing resources Bebop and Swing, clusters operated by the Laboratory Computing Resource Center, and Crux, operated by the Argonne Leadership Computing Facility, at Argonne National Laboratory.

The authors thank Prof. Michael Maire (University of Chicago) for helpful discussions and guidance on relevant machine learning approaches for generative SAR despeckling.

\appendices
\section{Hyperparameter Optimization}

\begin{table*}[!t]
    \centering
    \caption{Hyperparameters tuned for UNet and UNet++ multispectral and SAR models using Bayesian optimization.}
    \label{tab:s2_hyperparams}
    \begin{tabular}{l l l l}
        \toprule
        \textbf{Model} & \textbf{Hyperparameter} & \textbf{Type} & \textbf{Search Space} \\
        \midrule
        \multirow{4}{*}{UNet}  & Learning Rate & Continuous & $[1e^{-5}, 1e^{-2}]$ \\
                                & LR Scheduler & Discrete & $\{\text{Constant}, \text{ReduceLROnPlateau}\}$ \\
                                & Dropout Rate & Continuous & $[0.05, 0.50]$ \\
                                & Loss & Discrete & $\{\text{BCELoss}, \text{BCEDiceLoss}, \text{TverskyLoss}, \text{FocalTverskyLoss}\}$ \\
        \midrule
        \multirow{4}{*}{UNet++} & Learning Rate & Continuous & $[1e^{-5}, 1e^{-2}]$ \\
                                & LR Scheduler & Discrete & $\{\text{Constant}, \text{ReduceLROnPlateau}\}$ \\
                                & Dropout Rate & Continuous & $[0.05, 0.50]$ \\
                                & Loss & Discrete & $\{\text{BCELoss}, \text{BCEDiceLoss}, \text{TverskyLoss},
                                \text{FocalTverskyLoss}\}$ \\
                                & Deep Supervision & Discrete & $\{\text{True}, \text{False}\}$ \\
        \bottomrule
    \end{tabular}
\end{table*}

\section{Threshold Calibrated Model Performance}

\begin{table*}[!t]
\centering
\caption{Test metrics for multispectral and SAR models on all channels (except DEM) using the validation-set F1 optimal threshold.}
\label{tab:calibrated_threshold}
\begin{tabular}{l l c c c c}
\toprule
\textbf{Dataset} & \textbf{Model} & \textbf{Threshold} & \textbf{F1} & \textbf{Recall} & \textbf{Precision} \\
\midrule
\multirow{2}{*}{S2}
& UNet++ & 0.75 & 0.8998 & 0.9112 & 0.8886 \\
& UNet & 0.20 & 0.8967 & 0.9207 & 0.8740 \\
\midrule
\multirow{2}{*}{S1}
& UNet++ & 0.85 & 0.7532 & 0.6361 & 0.9235 \\
& UNet & 0.88 & 0.7546 & 0.6387 & 0.9220 \\
\bottomrule
\end{tabular}
\end{table*}

\section{Class-wise Performance Benchmarks}

\begin{table*}[!t]
    \centering
    \caption{NLCD (Urban) and SCL (Cloud Shadow, Topographic Shadow, Cirrus) class benchmarks for UNet++ S2 and S1 models using all channels (except DEM). Standard error is reported for mean overall F1 ($t = 0.5$) across $N=10$ trials. Note that the metrics for the S1 model are calculated using the shifted label with lowest model loss, not necessarily the central window.}
    \label{tab:class_f1}
    \begin{tabular}{l c c c c c}
    \toprule
    \textbf{Dataset} & \textbf{Model} & \multicolumn{4}{c}{\textbf{F1 ($t = 0.5$)}} \\
    \cmidrule(lr){3-6}
     &  & \textbf{Urban} & \textbf{Cloud Shadow} & \textbf{Topographic Shadow} & \textbf{Cirrus} \\
    \midrule
    S2 & UNet++ & $0.7871 \pm 0.0039$ & $0.7810 \pm 0.0064$ & $0.8767 \pm 0.0038$ & $0.8772 \pm 0.0086$ \\
    S1 & UNet++ & $0.3308 \pm 0.0087$ & $0.7171 \pm 0.0041$ & $0.8330 \pm 0.0009$ & $0.7551 \pm 0.0049$ \\
    \bottomrule
    \end{tabular}
\end{table*}

\section{Additional Qualitative Results}

\section*{Code and Data Availability}
The code used to generate the results in this paper is publicly available at
\url{https://github.com/davdma/floodmaps}. A subset of the data is available at
\url{https://zenodo.org/records/18528354}.

\bibliographystyle{IEEEtran}
\bibliography{IEEEabrv}

\newpage
 
\vspace{11pt}

\begin{IEEEbiography}[{\includegraphics[width=1in,height=1.25in,clip,keepaspectratio]{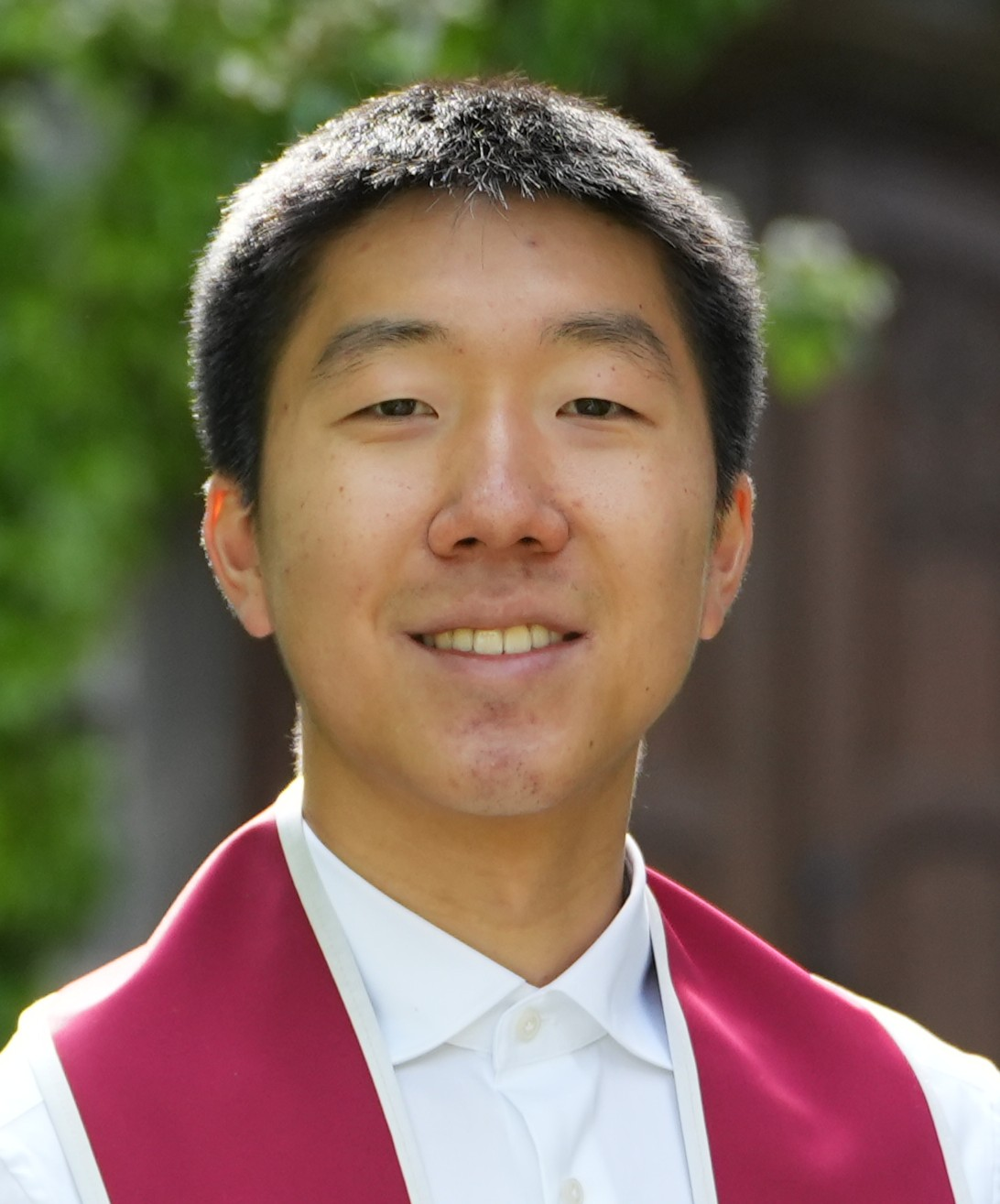}}]{David Ma}
received the B.S. degree in computer science and mathematics from the University of Chicago, Chicago, IL, USA, in 2025. He is currently a researcher in the Environmental Science Division at Argonne National Laboratory, Lemont, IL, USA, where he works on remote sensing and machine learning for high resolution flood mapping. His interests include high-performance computing, software systems infrastructure, performance optimization, and large-scale data processing pipelines.
\end{IEEEbiography}

\begin{IEEEbiographynophoto}{Jeremy Feinstein} is a research data scientist at Argonne National Laboratory in the Environmental Science Division, Department of Hydrology, Remediation, and Risk-Based Restoration, working on machine learning applications in hydrology and Earth system sciences, with applications to extreme value analysis, bioenergy, and flood prediction. His interests include remote sensing, information and reasoning systems, and scalable high-performance computing workflows.
\end{IEEEbiographynophoto}

\begin{IEEEbiography}[{\includegraphics[width=1in,height=1.25in,clip,keepaspectratio]{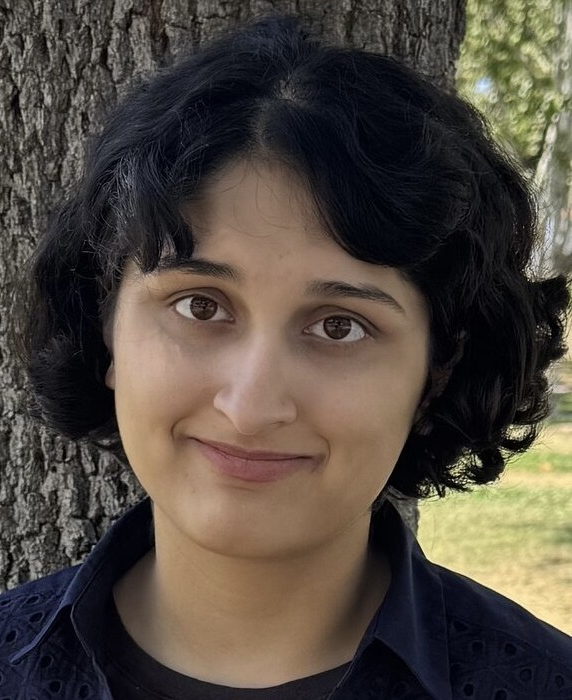}}]{Shreya Pandit}
is pursuing the B.S. degree in civil engineering. She is currently a student at Ohlone College, Fremont, CA, USA. She was a research intern in the Environmental Science Division at Argonne National Laboratory in 2025, working on data annotation, flood event collection, and evaluating transfer learning strategies for flood mapping.
\end{IEEEbiography}

\begin{IEEEbiography}[{\includegraphics[width=1in,height=1.25in,clip,keepaspectratio]{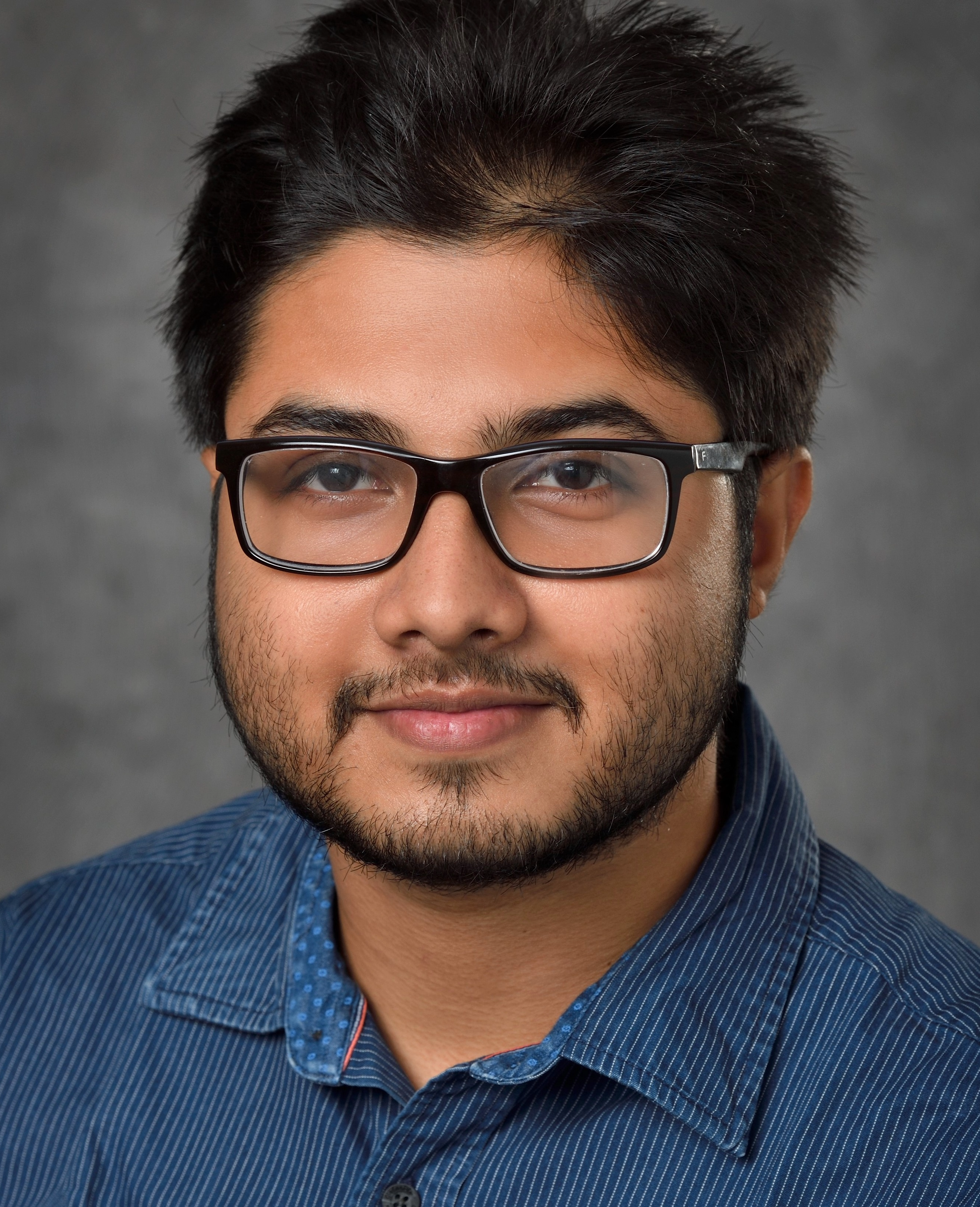}}]{Arkaprabha Ganguli} received his Ph.D. in Statistics from Michigan State University. Currently, he is a postdoctoral researcher at Argonne National Laboratory in the Mathematics and Computer Science Division. He also holds Bachelor’s and Master’s degrees in Statistics from the University of Calcutta, India.  His research interests center on statistical machine learning, deep generative models, high-dimensional data analysis, and uncertainty-aware modeling.
\end{IEEEbiography}

\begin{IEEEbiography}[{\includegraphics[width=1in,height=1.25in,clip,keepaspectratio]{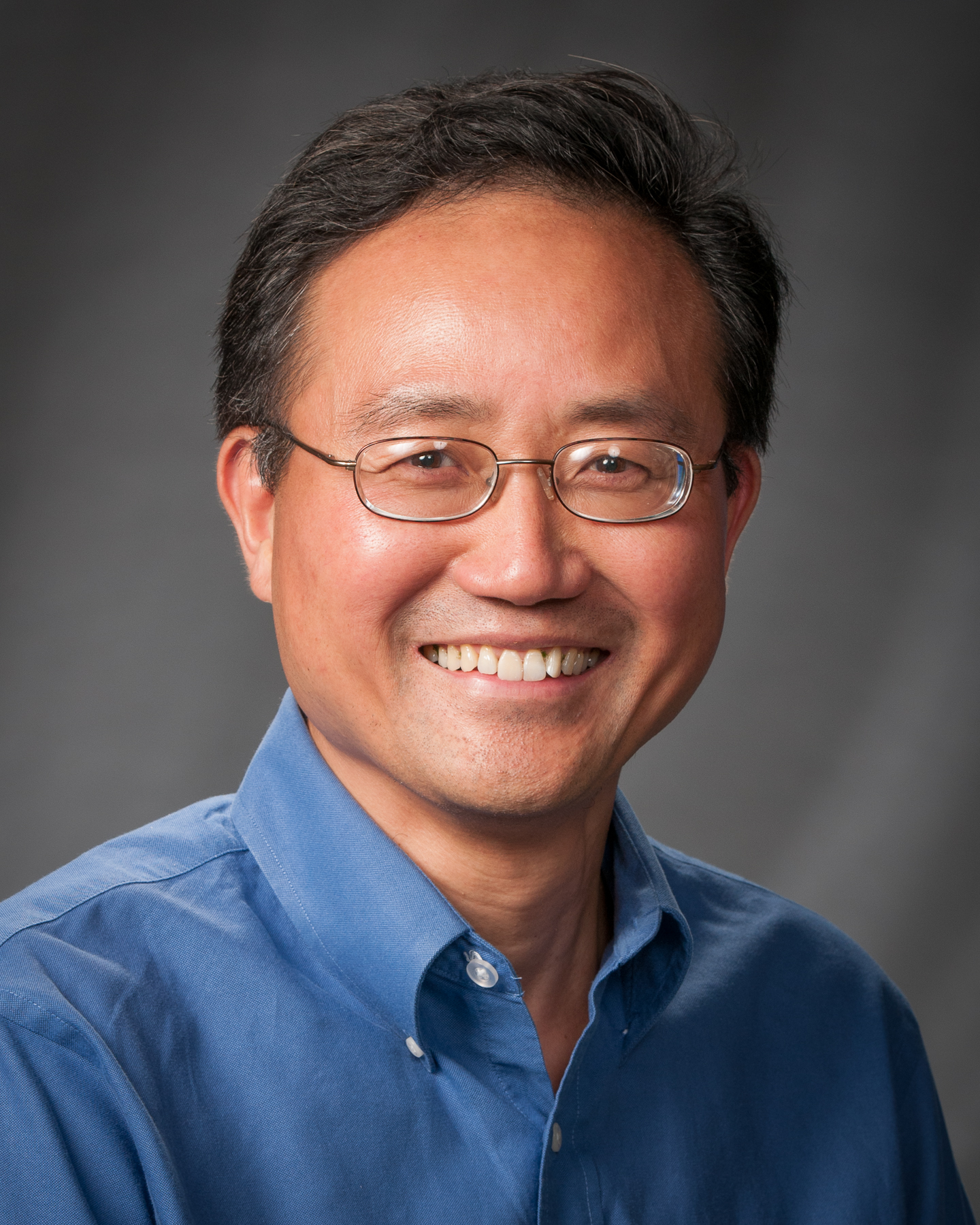}}]{Eugene Yan}received his Ph.D. in hydrogeology at Ohio State University. He currently serves as Department Head of Hydrology, Remediation, and Risk-based Restoration at Argonne National Laboratory and a Senior Fellow of Northwestern-Argonne Institute of Science and Engineering (NAISE). His research spans flood forecasting, modeling of hydrologic processes and contaminant fate and transport, groundwater remediation, and machine learning applications for complex hydrologic systems. 
\end{IEEEbiography}

\vspace{11pt}


\vfill

\end{document}